\documentclass{article} 
\usepackage{iclr2020_conference,times}

\usepackage{amsmath,amsfonts,bm}

\def\eqref#1{equation~\ref{#1}}

\def\1{\bm{1}}

\def\ra{{\textnormal{a}}}

\def\rr{{\textnormal{r}}}

\def\rx{{\textnormal{x}}}

\def\rva{{\mathbf{a}}}

\def\erva{{\textnormal{a}}}

\def\ervx{{\textnormal{x}}}

\def\rmA{{\mathbf{A}}}

\def\vmu{{\bm{\mu}}}
\def\vtheta{{\bm{\theta}}}
\def\va{{\bm{a}}}
\def\vb{{\bm{b}}}
\def\vc{{\bm{c}}}
\def\vd{{\bm{d}}}
\def\ve{{\bm{e}}}

\def\vu{{\bm{u}}}
\def\vv{{\bm{v}}}

\def\vx{{\bm{x}}}
\def\vy{{\bm{y}}}
\def\vz{{\bm{z}}}

\def\eva{{a}}

\def\mA{{\bm{A}}}
\def\mB{{\bm{B}}}
\def\mC{{\bm{C}}}
\def\mD{{\bm{D}}}
\def\mE{{\bm{E}}}

\def\mH{{\bm{H}}}
\def\mI{{\bm{I}}}
\def\mJ{{\bm{J}}}

\def\mL{{\bm{L}}}
\def\mM{{\bm{M}}}
\def\mN{{\bm{N}}}

\def\mP{{\bm{P}}}
\def\mQ{{\bm{Q}}}

\def\mS{{\bm{S}}}

\def\mU{{\bm{U}}}
\def\mV{{\bm{V}}}

\def\mX{{\bm{X}}}

\def\mSigma{{\bm{\Sigma}}}

\DeclareMathAlphabet{\mathsfit}{\encodingdefault}{\sfdefault}{m}{sl}
\SetMathAlphabet{\mathsfit}{bold}{\encodingdefault}{\sfdefault}{bx}{n}
\newcommand{\tens}[1]{\bm{\mathsfit{#1}}}
\def\tA{{\tens{A}}}

\def\tX{{\tens{X}}}

\def\gG{{\mathcal{G}}}

\def\sA{{\mathbb{A}}}
\def\sB{{\mathbb{B}}}

\def\sS{{\mathbb{S}}}

\def\emA{{A}}

\newcommand{\etens}[1]{\mathsfit{#1}}

\def\etA{{\etens{A}}}

\newcommand{\E}{\mathbb{E}}

\newcommand{\R}{\mathbb{R}}

\newcommand{\KL}{D_{\mathrm{KL}}}
\newcommand{\Var}{\mathrm{Var}}

\newcommand{\Cov}{\mathrm{Cov}}

\newcommand{\normltwo}{L^2}
\newcommand{\normlp}{L^p}

\newcommand{\parents}{Pa} 

\usepackage{url}
\usepackage[T1]{fontenc}    
\usepackage{url}            
\usepackage{booktabs}       
\usepackage{amsfonts}       
\usepackage{nicefrac}       
\usepackage{microtype}      
\usepackage{subfiles}
\usepackage{nicefrac}       
\usepackage{microtype}      
\usepackage{lipsum}
\usepackage{amsmath}
\usepackage{amssymb}
\usepackage{dsfont}
\usepackage{amsthm}
\usepackage{graphicx}
\usepackage{xcolor}
\usepackage{subcaption}
\usepackage{hyperref}
\hypersetup{colorlinks,linkcolor={blue},citecolor={blue},urlcolor={blue}} 
\usepackage{caption}
\usepackage{cleveref}
\usepackage{subfiles}
\usepackage{algorithm2e}
\iclrfinalcopy
\usepackage{bm}
\usepackage{thm-restate}

\newtheorem{lemma}{Lemma}
\newtheorem{theorem}{Theorem}
\newtheorem{corollary}{Corollary}

\numberwithin{equation}{section}
\numberwithin{theorem}{section}
\numberwithin{lemma}{section}
\numberwithin{corollary}{section}

\newcommand{\red}[1]{{#1}}
\newcommand{\zero}{\mathbf{0}}

\newcommand{\bb}[1]{\textbf{#1}}
\newcommand{\x}[1]{\vx^{(#1)}}
\newcommand{\y}[1]{\vy^{(#1)}}
\newcommand{\z}[1]{\vz^{(#1)}}
\def \a {\alpha}
\def \b {\beta}
\def \R {\mathbb{R}}
\def \E {\mathds{E}}
\def \N {\mathcal{N}}
\def \s {\sigma}
\def \d {\delta}
\def \bmat {\begin{matrix}}
\def \emat {\end{matrix}}
\def \l {\lambda}

\def \be {\begin{eqnarray}}
\def \en {\end{eqnarray}}
\def \gr {\nabla}
\def \tr {\nonumber \\}
\def \rr {\mathcal{R}}

\def \k {\kappa}
\def \n {\nabla}

\def \vphi {\bm \phi}
\newcommand{\blue}[1]{\textcolor{blue}{#1}}

\title{Convergence of Gradient Methods on Bilinear Zero-Sum Games}

\author{Guojun Zhang \& Yaoliang Yu 
\\ School of Computer Science\\
University of Waterloo\\
Vector Institute\\
\texttt{\{guojun.zhang,yaoliang.yu\}@uwaterloo.ca} \\
}

\begin{document}

\maketitle

\begin{abstract}
Min-max formulations have attracted great attention in the ML community due to the rise of deep generative models and adversarial methods, while understanding the dynamics of gradient algorithms for solving such formulations has remained a grand challenge. As a first step, we restrict to bilinear zero-sum games and give a systematic analysis of popular gradient updates, for both simultaneous and alternating versions. We provide exact conditions for their convergence and find the optimal parameter setup and convergence rates. In particular, our results offer formal evidence that alternating updates converge ``better'' than simultaneous ones.
\end{abstract}

\section{Introduction}\label{intro}

 Min-max optimization has received significant attention recently due to the popularity of generative adversarial networks (GANs) \citep{goodfellow2014generative}, adversarial training \citep{madry2017towards} and reinforcement learning \citep{du2017stochastic, dai2018sbeed}, just to name some examples. Formally, given a bivariate function $f(\displaystyle \vx, \vy)$,
 we aim to find a \emph{saddle point} $(\displaystyle \vx^*, \displaystyle \vy^*)$ such that
\begin{align}
\label{eq:saddle}
f(\displaystyle \vx^*, \displaystyle \vy) \leq f(\displaystyle \vx^*, \displaystyle \vy^*) \leq f(\displaystyle \vx, \displaystyle \vy^*), \, \forall \displaystyle \vx\in \mathbb{R}^n, \, \forall \displaystyle \vy\in \mathbb{R}^n.
\end{align}
Since the beginning of game theory, various algorithms have been proposed for finding saddle points \citep{arrow1958studies,DemyanovPevnyi72,Goldstein72,korpelevich1976extragradient,rockafellar1976monotone,Bruck77,Lions78,nemirovsky1983problem,freund1999adaptive}. Due to its recent resurgence in ML, new algorithms specifically designed for training GANs were proposed \citep{daskalakis2017training, kingma2014adam, GidelHPHLLM19, mescheder2017numerics}.
However, due to the inherent non-convexity in deep learning formulations, our current understanding of the convergence behaviour of new and classic gradient algorithms 
is still quite limited, and existing analysis mostly focused on bilinear games or strongly-convex-strongly-concave games \citep{tseng1995linear, daskalakis2017training, GidelHPHLLM19, LiangStokes18, mokhtari2019unified}.
Non-zero-sum bilinear games, on the other hand, are known to be PPAD-complete \citep{chen2009settling} (for finding approximate Nash equilibria, see e.g. \cite{deligkas2017computing}). 

In this work, we study bilinear zero-sum games as a first step towards understanding general min-max optimization, although our results apply to some simple GAN settings \citep{gidel2018variational}. It is well-known that certain gradient algorithms converge linearly on bilinear zero-sum games \citep{LiangStokes18, mokhtari2019unified, rockafellar1976monotone, korpelevich1976extragradient}. These iterative algorithms usually come with two versions: \emph{Jacobi} style updates or \emph{Gauss--Seidel} (GS) style.
In a Jacobi style, we update the two sets of parameters (i.e., $\displaystyle \vx$ and $\displaystyle \vy$) \textit{simultaneously} whereas in a GS style we update them  \textit{alternatingly} (i.e., one after the other). Thus, Jacobi style updates are naturally amenable to parallelization while GS style updates have to be sequential, although the latter is usually found to converge faster (and more stable).
In numerical linear algebra, the celebrated Stein--Rosenberg theorem \citep{stein1948solution} formally proves that in solving certain linear systems, GS updates converge \emph{strictly} faster than their Jacobi counterparts, and often with a larger set of convergent instances. However, this  result does not readily apply to bilinear zero-sum games.

Our main goal here is to answer the following questions about solving bilinear zero-sum games:
\begin{itemize}
\item When exactly does a gradient-type algorithm converge?
\item What is the optimal convergence rate by tuning the step size or other parameters?
\item Can we prove something similar to the Stein--Rosenberg theorem for Jacobi and GS updates?
\end{itemize}
\paragraph{Contributions} We summarize our main results from \S\ref{schur_ana} and \S\ref{opt_spe} in Table \ref{j_vs_gs} and \ref{opt_rate} respectively, with supporting experiments given in \S\ref{sec:exp}. We use $\sigma_1$ and $\sigma_n$ to denote the largest and the smallest singular values of matrix $\mE$ (see~\eqref{bilinear_c}), and $\kappa := \sigma_1/\sigma_n$ denotes the condition number. The algorithms will be introduced in \S\ref{prem}. Note that we generalize gradient-type algorithms but retain the same names. Table \ref{j_vs_gs} shows that in most cases that we study, whenever Jacobi updates converge, the corresponding GS updates converge as well (usually with a faster rate), but the converse is not true (\S 3). This extends the well-known Stein--Rosenberg theorem to bilinear games. Furthermore, Table \ref{opt_rate} tells us that by generalizing existing gradient algorithms, we can obtain faster convergence rates. 

\begin{table*}
  \caption{Comparisons between Jacobi and Gauss--Seidel updates. The second and third columns show when exactly an algorithm converges, with Jacobi or GS updates. The last column shows whether the convergence region of Jacobi updates is contained in the GS convergence region. }
  \label{j_vs_gs}
  \centering
  \begin{tabular}{llll}
    \toprule
    Algorithm  &  Jacobi & Gauss--Seidel & Contained? \\
    \midrule
     GD  & diverges & limit cycle & N/A \\
     EG & Theorem \ref{thm_eg} & Theorem \ref{thm_eg} & if $\b_1 + \b_2 + \a^2 < 2/\s_1^2$ \\
     OGD & Theorem \ref{schur_ogd} & Theorem \ref{schur_ogd}  & yes \\
     momentum & does not converge & Theorem \ref{thm_momentum} & yes \\
    \bottomrule
  \end{tabular}
\end{table*}

\begin{table*}
  \caption{Optimal convergence rates. In the second column, $\beta_*$ denotes a specific parameter that depends on $\sigma_1$ and $\sigma_n$ (see~\eqref{best_b}). In the third column, the linear rates are for large $\kappa$. The optimal parameters for both Jacobi and Gauss--Seidel EG algorithms are the same. \red{$\a$ denotes the step size ($\a_1 = \a_2 = \a$), and $\b_1$ and $\b_2$ are hyper-parameters for EG and OGD, as given in \S\ref{prem}.} }
  \label{opt_rate}
  \centering
  \begin{tabular}{llllll}
    \toprule
    Algorithm  & $\a$ & $\b_1$ & $\b_2$ & Rate exponent & Comment \\
    \midrule 
    EG & $\sim 0$ & $2/(\s_1^2 + \s_n^2)$ & $\b_1$ & $\sim 1 - 2/\kappa^2$ & Jacobi and Gauss--Seidel\\
    Jacobi OGD & $2\b_1$ & $\b_*$ & $\b_1$ & $\sim 1 - 1/(6\kappa^2)$ & $\b_1 = \b_2 = \a/2$ \\
    GS OGD & $\sqrt{2}/\s_1$ & $\sqrt{2}\s_1/(\s_1^2 + \s_n^2)$ & $0$ & $\sim  1 - 1/\kappa^2$ & $\b_1$ and $\b_2$ can interchange\\
   GS momentum & $\a_*$ & $-1/2$ & $0$ & $\sim 1 - 2/(9\kappa^2)$ & \\
    \bottomrule
  \end{tabular}
\end{table*}



\section{Preliminaries}\label{prem}
\red{In the study of GAN training, bilinear games are often regarded as an important simple example for theoretically analyzing and  understanding new algorithms and techniques \citep[e.g.][]{daskalakis2017training, gidel2018variational, GidelHPHLLM19, LiangStokes18}. It captures the difficulty in GAN training and can represent some simple GAN formulations \citep{arjovsky2017wasserstein, daskalakis2017training, gidel2018variational, mescheder2018training}.} Mathematically, \textit{bilinear} zero-sum games can be formulated as the following min-max problem:
 \begin{equation}\label{bilinear_c}
 \textstyle
     \min_{ \vx\in \R^n} \max_{ \vy \in \R^n} ~~\vx^\top \mE \vy + \vb^\top \vx + \vc^\top \vy.
 \end{equation}
The set of all saddle points (see definition in \cref{eq:saddle}) is:
\begin{equation}\label{sp}
   \{(\vx, \vy) \,| \, \mE\vy + \vb = {\bf 0}, \, \mE^\top \vx + \vc = {\bf 0}\}.
\end{equation}
Throughout, for simplicity we assume $\mE$ to be invertible, \red{whereas the seemingly general case with non-invertible $\mE$ is treated in Appendix \ref{app:singular}.} The linear terms are not essential in our analysis and we take $\vb = \vc = {\bf 0}$ throughout the paper\footnote{If they are not zero, one can translate $\vx$ and $\vy$ to cancel the linear terms, \red{see e.g.~\citet{GidelHPHLLM19}.}}. \red{In this case, the only saddle point is $({\bf 0}, {\bf 0})$.}
For \red{bilinear} games, it is well-known that simultaneous gradient descent ascent does not converge \citep{nemirovsky1983problem} and other gradient-based algorithms tailored for min-max optimization have been proposed  \citep{korpelevich1976extragradient,daskalakis2017training,gidel2018variational,mescheder2017numerics}. These iterative algorithms all belong to the class of general linear dynamical systems (LDS, a.k.a. matrix iterative processes). Using state augmentation $\displaystyle \vz^{(t)} :=(\vx^{(t)}, \vy^{(t)})$ we define a general $k$-step LDS as follows:
\begin{align}
\label{eq:LDS}
\textstyle
\vz^{(t)} = \sum_{i=1}^k \mA_i \vz^{(t-i)} + \vd,
\end{align}
where the matrices $\mA_i$ and vector $\vd$ depend on the gradient algorithm (examples can be found in Appendix~\ref{appe:char_poly}).
Define the characteristic polynomial\red{, with $\mA_0 = -\mI$:
\begin{align}\label{sec2.2: char_poly}
\textstyle
    p(\lambda) := \det( \sum_{i=0}^k \mA_i \lambda^{k-i}).
\end{align}}
The following well-known result decides when such a $k$-step LDS converges for any initialization:
\begin{theorem}[{e.g. \cite{GohbergLR82}}]
\label{thm:spec}
The LDS in \cref{eq:LDS} converges for any initialization $(\vz^{(0)}, \ldots, \vz^{(k-1)})$ iff the spectral radius $r := \max \{|\l|: p(\l) = 0\} < 1$, in which case $\{\vz^{(t)}\}$ converges linearly with an (asymptotic) exponent $r$.
\end{theorem}

Therefore, understanding the bilinear game dynamics reduces to spectral analysis. The (sufficient and necessary) convergence condition reduces to that all roots of $p(\l)$ lie in the (open) unit disk, which can be conveniently analyzed through the celebrated Schur's theorem \citep{schur1917potenzreihen}:
\begin{theorem}[\citet{schur1917potenzreihen}]\label{schur}
The roots of a real polynomial $p(\l) = a_0 \l^n + a_1 \l^{n-1} + \dots + a_n$ are within the (open) unit disk of the complex plane iff $\forall k \in \{1, 2, \dots, n\}, ~ \det(\mP_k \mP_k^{\top} - \mQ_k^{\top}\mQ_k) > 0$, where $\mP_k, \mQ_k$ are $k\times k$ matrices defined as: $[\mP_k]_{i,j} = a_{i-j}\1_{i\geq j}$, $[\mQ_k]_{i,j} = a_{n-i+j}\1_{i\leq j}$. 

\end{theorem}

In the theorem above, we denoted $\1_{S}$ as the indicator function of the event $S$, i.e. $\1_S = 1$ if $S$ holds and $\1_S=0$ otherwise.
For a nice summary of related stability tests, see \citet{Mansour11}. We therefore define \textit{Schur stable} polynomials to be those polynomials whose roots all lie within the (open) unit disk of the complex plane. Schur's theorem has the following corollary (proof included in Appendix \ref{proof_corollary} for the sake of completeness):
\begin{restatable}[e.g. \cite{Mansour11}]{corollary}{quadpoly}\label{schur_234}
A real quadratic polynomial $\l^2 + a \l + b$ is Schur stable iff $b < 1, \, |a| < 1 + b$; A real cubic polynomial $\l^3 + a\l^2 + b \l + c$ is Schur stable iff $|c| < 1$, $|a+c|<1+b$, $b - a c < 1- c^2$; A real quartic polynomial $\l^4 + a \l^3 + b \l^2 + c\l + d$ is Schur stable iff $|c - ad| < 1 - d^2$, $|a+c| < b+d+1$, and $b < (1+d) + (c- a d)(a-c)/(d-1)^2$.
\end{restatable}

Let us formally define Jacobi and GS updates: Jacobi updates take the form 
\begin{align*}
    \vx^{(t)} = T_1(\vx^{(t-1)}, \vy^{(t-1)}, \ldots, \vx^{(t-k)}, \vy^{(t-k)}), \,
    \vy^{(t)} = T_2(\vx^{(t-1)}, \vy^{(t-1)}, \ldots, \vx^{(t-k)}, \vy^{(t-k)}),
\end{align*}
while Gauss--Seidel updates replace $\vx^{(t-i)}$ with the more recent $\vx^{(t-i+1)}$ in operator $T_2$, where $T_1, T_2: \R^{nk}\times \R^{nk} \to \R^n$ can be any update functions. For LDS updates in \cref{eq:LDS} we find a nice relation between the characteristic polynomials of Jacobi and GS updates in  \Cref{thm:JacobGS} (proof in \Cref{pf:JacobiGS}), which turns out to greatly simplify our subsequent analyses:
\begin{restatable}[\textbf{Jacobi vs. Gauss--Seidel}]{theorem}{JGS}{}
\label{thm:JacobGS}
\red{Let $p(\lambda, \gamma) = \det(\sum_{i=0}^{k} (\gamma \mL_i + \mU_i) \lambda^{k-i})$,} where $\mA_i = \mL_i + \mU_i$ and $\mL_i$ is strictly lower block triangular. Then, the characteristic polynomial of Jacobi updates is $p(\lambda, 1)$ while that of Gauss--Seidel updates is $p(\lambda, \lambda)$.
\end{restatable}
Compared to the Jacobi update, in some sense the Gauss--Seidel update amounts to \emph{shifting the strictly lower block triangular matrices $\mL_i$ one step to the left}, as $p(\l, \l)$ can be rewritten as \red{$\det\left( \sum_{i=0}^{k} (\mL_{i+1} + \mU_i) \lambda^{k-i}\right)$}, with $\mL_{k+1} := {\bf 0}$. This observation will significantly simplify our comparison between Jacobi and Gauss--Seidel updates. 

Next, we define some popular gradient algorithms for finding saddle points in the min-max problem 
\begin{align}
\min_{\vx} \max_{\vy} f(\vx, \vy).
\end{align}
We present the algorithms for a general (bivariate) function $f$ although our main results will specialize $f$ to the bilinear case in \cref{bilinear_c}.
Note that we introduced more ``step sizes'' for our refined analysis, as we find that the enlarged parameter space often contains choices for faster linear convergence (see \S\ref{opt_spe}). We only define the Jacobi updates, while the GS counterparts can be easily inferred. We always use $\a_1$ and $\a_2$ to define step sizes (or learning rates) which are positive.

\paragraph{\textbf{Gradient descent (GD)}} The generalized GD update has the following form:
\be
\label{g_gd}\vx^{(t+1)} = \vx^{(t)} - \a_1 \gr_{\vx} f(\vx^{(t)}, \vy^{(t)}), \qquad \vy^{(t+1)} = \vy^{(t)} + \a_2 \gr_{\vy} f(\vx^{(t)}, \vy^{(t)}).
\en
When $\a_1 = \a_2$, the convergence of averaged iterates (a.k.a. Cesari convergence) for convex-concave games is analyzed in \citep{Bruck77,nemirovski1978cesari, nedic2009subgradient}. Recent progress on interpreting GD with dynamical systems can be seen in, e.g.,  \cite{mertikopoulos2018cycles, bailey2019finite, bailey2018multiplicative}.

\paragraph{\textbf{Extra-gradient (EG)}} We study a generalized version of EG, defined as follows: 
\begin{align}
\label{g_eg_1}&\!\!\vx^{(t+1/2)} = \vx^{(t)} - \gamma_1 \gr_{\vx} f(\vx^{(t)}, \vy^{(t)}), \, \vy^{(t+1/2)} = \vy^{(t)} + \gamma_2 \gr_{\vy} f(\vx^{(t)}, \vy^{(t)}); \\ 
\label{g_eg_2}&\!\!\vx^{(t+1)} = \vx^{(t)} - \a_1 \gr_{\vx} f(\vx^{(t+1/2)}, \vy^{(t+1/2)}), \,  \vy^{(t+1)} = \vy^{(t)} + \a_2 \gr_{\vy} f(\vx^{(t+1/2)}, \vy^{(t+1/2)}).
\end{align}
EG was first proposed in \citet{korpelevich1976extragradient} with the restriction $\a_1 = \a_2 = \gamma_1 = \gamma_2$, under which linear convergence was proved for bilinear games. Convergence of EG on convex-concave games was analyzed in \citet{nemirovski2004prox, monteiro2010complexity}, and \cite{mertikopoulos2019optimistic} provides convergence guarantees for specific non-convex-non-concave problems. For bilinear games, a slightly more generalized version was proposed in \citet{LiangStokes18} where $\a_1 = \a_2$, $\gamma_1 = \gamma_2$, with linear convergence proved. For later convenience we define $\b_1 = \a_2 \gamma_1$ and $\b_2 = \a_1 \gamma_2$. 

\paragraph{\textbf{Optimistic gradient descent (OGD)}} We study a generalized version of OGD, defined as follows: 
\begin{align}
\label{g_ogd_1}\vx^{(t+1)} &= \vx^{(t)} - \a_1 \gr_{\vx} f(\vx^{(t)}, \vy^{(t)}) + \b_1 \gr_{\vx} f(\vx^{(t-1)}, \vy^{(t-1)}), \\
\label{g_ogd_2}\vy^{(t+1)} &= \vy^{(t)} + \a_2 \gr_{\vy} f(\vx^{(t)}, \vy^{(t)}) - \b_2 \gr_{\vy} f(\vx^{(t-1)}, \vy^{(t-1)}).
\end{align}
The original version of OGD was given in \cite{popov1980modification} with $\a_1 = \a_2 = 2\b_1 = 2\b_2$ and rediscovered in the GAN literature \citep{daskalakis2017training}. Its linear convergence for bilinear games was proved in \citet{LiangStokes18}. A slightly more generalized version with $\a_1 = \a_2$ and $\b_1 = \b_2$ was analyzed in \citet{peng2019training, mokhtari2019unified}, again with linear convergence proved. The stochastic case was analyzed in \cite{hsieh2019convergence}.

\paragraph{\textbf{Momentum method}} Generalized heavy ball method was analyzed in \citet{GidelHPHLLM19}: 
\begin{align}
\label{g_hb_1}\vx^{(t+1)} &= \vx^{(t)} - \a_1 \gr_{\vx} f(\vx^{(t)}, \vy^{(t)}) + \b_1 ( \vx^{(t)} -  \vx^{(t-1)}),\\
\label{g_hb_2}\vy^{(t+1)} &= \vy^{(t)} + \a_2 \gr_{\vy} f(\vx^{(t)}, \vy^{(t)}) + \b_2 ( \vy^{(t)} -  \vy^{(t-1)}).
\end{align}
This is a modification of Polyak's heavy ball (HB) \citep{Polyak64}, which also motivated Nesterov's accelerated gradient algorithm (NAG) \citep{nesterov1983method}.
Note that for both $\vx$-update and the $\vy$-update, we \emph{add} a scale multiple of the successive difference (e.g. proxy of the momentum). For this algorithm our result below improves those obtained in \citet{GidelHPHLLM19}, as will be discussed in \S\ref{schur_ana}.

\paragraph{EG and OGD as approximations of proximal point algorithm} It has been observed recently in \cite{mokhtari2019unified} that for convex-concave games, EG ($\a_1 = \a_2 = \gamma_1 = \gamma_2 = \eta$) and OGD ($\a_1/2 = \a_2/2 = \b_1 = \b_2 = \eta$) can be treated as approximations of the proximal point algorithm \citep{Martinet70, rockafellar1976monotone} when $\eta$ is small. With this result, one can show that EG and OGD converge to saddle points sublinearly for smooth convex-concave games \citep{mokhtari2019proximal}. We give a brief introduction of the proximal point algorithm in  \Cref{app:pp} (including a linear convergence result for the slightly generalized version).

The above algorithms, when specialized to a bilinear function $f$ (see~\cref{bilinear_c}), can be rewritten as a 1-step or 2-step LDS (see.~\cref{eq:LDS}). See  ~\Cref{appe:char_poly} for details.


\section{Exact conditions}\label{schur_ana}
With tools from \S\ref{prem}, we formulate necessary and sufficient conditions under which a gradient-based algorithm converges for bilinear games. We sometimes use ``J'' as a shorthand for Jacobi style updates and ``GS'' for Gauss--Seidel style updates. For each algorithm, we first write down the characteristic polynomials (see derivation in  \Cref{appe:char_poly}) for both Jacobi and GS updates, and present the exact conditions for convergence. Specifically, we show that in many cases the GS convergence regions strictly include the Jacobi convergence regions. The proofs for  \Cref{thm-gd}, \ref{thm_eg}, \ref{schur_ogd} and \ref{thm_momentum} can be found in \Cref{appen:schur_gd}, \ref{appen:schur_eg}, \ref{appen:schur_ogd}, and \ref{append:pf_momen}, respectively. 
\vspace{-0.2em}
\paragraph{GD} The characteristic equations can be computed as:
\begin{eqnarray}
\label{gd_j} \textrm{J:}\;(\l - 1)^2 + \a_1 \a_2 \sigma^2 = 0,\, \textrm{GS:}\;(\l - 1)^2 + \a_1 \a_2 \sigma^2 \l = 0.
\end{eqnarray}
\paragraph{Scaling symmetry} From \cref{gd_j} we obtain a scaling symmetry $(\a_1, \a_2) \to (t\a_1, \a_2/t)$, with $t > 0$. With this symmetry we can always fix $\a_1 = \a_2 = \a$. This symmetry also holds for EG and momentum. For OGD, the scaling symmetry is slightly different with $(\a_1, \b_1, \a_2, \b_2) \to (t\a_1, t\b_1, \a_2/t, \b_2/t)$, but we can still use this symmetry to fix $\a_1 = \a_2 = \a$. 
\begin{restatable}[\textbf{GD}]{theorem}{GD}\label{thm-gd}
Jacobi GD and Gauss--Seidel GD do not converge. However, Gauss--Seidel GD can have a limit cycle while Jacobi GD always diverges. 
\end{restatable}
In the constrained case, \citet{mertikopoulos2018cycles} and \cite{bailey2018multiplicative} show that FTRL, a more generalized algorithm of GD, does not converge for polymatrix games. 
\red{When $\a_1 = \a_2$, the result of Gauss--Seidel GD has been shown in \citet{bailey2019finite}. }
\paragraph{EG} The characteristic equations can be computed as:
\be
   \label{eg_j}  \textrm{J:}&\; (\l - 1)^2 + (\b_1 + \b_2)\sigma^2(\l - 1) + (\a_1 \a_2 \s^2 + \b_1 \b_2 \s^4) = 0,\\
   \label{eg_gs}  \textrm{GS:}&\; (\l - 1)^2 + (\a_1 \a_2 + \b_1 + \b_2)\sigma^2(\l - 1) + (\a_1 \a_2 \s^2 + \b_1 \b_2 \s^4) = 0.
\en
\begin{restatable}
[\textbf{EG}]{theorem}{EG}\label{thm_eg}
For generalized EG with $\a_1 = \a_2 = \a$ and $\gamma_i = \b_i / \a$, Jacobi and Gauss--Seidel updates achieve linear convergence iff for any singular value $\s$ of $\mE$, we have:
\begin{eqnarray}
\label{eg_j_schur} &&\mathrm{J:}\,  |\b_1 \s^2 + \b_2 \s^2 - 2| < 1 + (1-\b_1 \s^2)(1-\b_2 \s^2)+ \a^2 \s^2, \tr &&(1-\b_1 \s^2 )(1 -\b_2 \s^2) + \a^2 \s^2 < 1,\\
\label{eg_gs_schur}&&\mathrm{GS:}\, |(\b_1 + \b_2 + \a^2)\s^2  - 2| < 1 + ( 1- \b_1 \s^2)(1- \b_2 \s^2),\tr
&& (1 - \b_1\s^2 )(1 - \b_2 \s^2)  < 1.
\end{eqnarray}
If $\b_1 + \b_2 + \a^2 < 2 /\s_1^2$, the convergence region of GS updates \textbf{strictly} include that of Jacobi updates. 
\end{restatable}
\paragraph{OGD} The characteristic equations can be computed as:
\begin{eqnarray}
   \label{ogd_j}\textrm{J:}&\;\l^2 (\l - 1)^2 + (\l \a_1 - \b_1)(\l \a_2 - \b_2)  \s^2 = 0,\\
   \label{ogd_gs}\textrm{GS:}&\;\l^2 (\l - 1)^2 + (\l \a_1 - \b_1)(\l \a_2 - \b_2) \l \s^2 = 0.
\end{eqnarray}

\begin{restatable}[\textbf{OGD}]
{theorem}{OGD}\label{schur_ogd}
For generalized OGD with $\a_1 = \a_2 = \a$, Jacobi and Gauss--Seidel updates achieve linear convergence iff for any singular value $\s$ of $\mE$, we have:
\begin{eqnarray}
\label{jacobi_schur}    \mathrm{J:}& \begin{cases}
    |\b_1 \b_2 \s^2| < 1, \, (\a- \b_1)(\a - \b_2) > 0, \, 4 + (\a + \b_1)(\a + \b_2) \s^2 > 0, \\
     \a^2 \left(\b_1^2 \s^2+1\right) \left(\b_2^2 \s^2+1\right)<({\beta_1} {\b_2} \s^2 +1) (2 \alpha  ({\b_1}+{\b_2})+{\b_1} {\b_2} ({\b_1} {\b_2} \s^2-3));
    \end{cases}\\
 \label{gs_schur}  \mathrm{GS:}&
   \begin{cases}
   (\a - \b_1)(\a - \b_2) > 0, (\alpha +{\b_1}) (\alpha +{\b_2}) \s^2 <4,\\
    (\alpha  {\b_1} \s^2 +1) (\alpha  {\b_2} \s^2 +1)>(1+\b_1 \b_2 \s^2 )^2.
    \end{cases}
\end{eqnarray}
The convergence region of GS updates \textbf{strictly} include that of Jacobi updates. 
\end{restatable}
\paragraph{Momentum} The characteristic equations can be computed as:
\begin{eqnarray}
\label{hb_j}&&\textrm{J: }(\l - 1)^2 (\l - \b_1)(\l - \b_2) + \a_1 \a_2 \s^2 \l^2 = 0,\\
\label{hb_gs}&&\textrm{GS: }(\l - 1)^2 (\l - \b_1)(\l - \b_2) + \a_1 \a_2 \s^2 \l^3 = 0.
\end{eqnarray}

\begin{restatable}[\textbf{momentum}]{theorem}{Momentum}\label{thm_momentum}
For the generalized momentum method with $\a_1 = \a_2 = \a$, the Jacobi updates never converge, while the GS updates converge iff for any singular value $ \s$ of $\mE$, we have:
\begin{eqnarray}\label{gs_momentum}
 &&| {\b_1} {\b_2}| <1,| -\alpha^2 \s^2 +{\b_1}+{\b_2}+2| <{\b_1} {\b_2}+3,\, 4 ({\b_1}+1) ({\b_2}+1)>\a^2 \s^2,\,\tr
  &&\a^2 \s^2 \b_1 \b_2 <(1-\b_1 \b_2)(2\b_1 \b_2 - \b_1 - \b_2).
\end{eqnarray}
This condition implies that at least one of $\b_1, \b_2$ is \textbf{negative}.
\end{restatable}

Prior to our work, only sufficient conditions for linear convergence were given for the usual EG and OGD; see~\S\ref{prem} above. For the momentum method, our result improves upon \citet{GidelHPHLLM19} where they only considered specific cases of parameters. For example, they only considered $\beta_1 = \b_2 \geq -1/16$ for Jacobi momentum \red{(but with explicit rate of divergence)}, and $\b_1 = -1/2$, $\b_2 = 0$ for GS momentum \red{(with convergence rate)}. Our  \Cref{thm_momentum} gives a more complete picture and formally justifies the necessity of negative momentum.

In the theorems above, we used the term ``convergence region'' to denote a subset of the parameter space (with parameters $\a$, $\b$ or $\gamma$) where the algorithm converges. Our result shares similarity with the celebrated Stein--Rosenberg theorem \citep{stein1948solution}, which only applies to solving linear systems with non-negative matrices (if one were to apply it to our case, the matrix $\mS$ in \cref{saddle_point_eq} in \Cref{split} needs to have non-zero diagonal entries, which is not possible).
In this sense, our results extend the Stein--Rosenberg theorem to cover nontrivial bilinear games.

\section{Optimal exponents of linear convergence}\label{opt_spe}

In this section we study the optimal convergence rates of EG, OGD and the momentum method. We define the exponent of linear convergence as $r = \lim_{t\to \infty} ||\vz^{(t)}||/||\vz^{(t - 1)}||$ which is the same as the spectral radius. For ease of presentation we fix $\a_1 = \a_2 = \a > 0$ (using scaling symmetry) and we use $r_*$ to denote the optimal exponent of linear convergence (achieved by tuning the parameters $\a, \beta, \gamma$). Our results show that by generalizing gradient algorithms one can obtain better convergence rates.
\begin{restatable}[\textbf{EG optimal}]{theorem}{EGopt}\label{eg_opt}
Both Jacobi and GS EG achieve the optimal exponent of linear convergence $r_* = (\k^2 - 1)/(\k^2 + 1)$ at $\a \to 0$ and $\b_1 = \b_2 = 2/(\s_1^2 + \s_n^2)$. As $\kappa \to \infty$, $r_* \to 1 - 2/\k^2$. 
\end{restatable}
Note that we defined $\b_i = \gamma_i \a$ in Section \ref{prem}. In other words, we are taking very large extra-gradient steps ($\gamma_i \to \infty$) and very small gradient steps ($\a\to 0$).


\begin{restatable}[\textbf{OGD optimal}]{theorem}{OGDopt}\label{ogd_opt}
For Jacobi OGD with $\b_1 = \b_2 = \b$, to achieve the optimal exponent of linear convergence, we must have $\a \leq 2\b$.  For the original OGD with $\a = 2\b$, the optimal exponent of linear convergence $r_*$ satisfies 
\be\label{best_r}
    r_*^2 = \frac{1}{2} + \frac{1}{4\sqrt{2}\s_1^2}\sqrt{(\s_1^2 - \s_n^2)(5 \s_1^2 - \s_n^2 + \sqrt{(\s_1^2 - \s_n^2)(9 \s_1^2 - \s_n^2)})}, \textrm{ at }
    \en
    \begin{equation}\label{best_b}
        \b_* = \frac{1}{4\sqrt{2}}\sqrt{\frac{3 \s_1^4 - (\s_1^2 - \s_n^2)^{3/2}\sqrt{9\s_1^2 - \s_n^2} + 6 \s_1^2 \s_n^2 - \s_n^4}{\s_1^4 \s_n^2}}.
    \end{equation}
    If $\k \to \infty$, $r_* \sim 1 - 1/(6\k^2)$. For GS OGD with $\b_2 = 0$, the optimal exponent of convergence is $r_* = \sqrt{(\k^2 - 1)/(\k^2 + 1)}$, at $\a = \sqrt{2}/\s_1$ and $\b_1 = \sqrt{2}\s_1/(\s_1^2 + \s_n^2)$. If $\k \to \infty$, $r_* \sim 1 - 1/\k^2$. 
\end{restatable}
\paragraph{Remark} The original OGD \citep{popov1980modification, daskalakis2017training} with $\a = 2\b$ may not always be optimal. For example, take one-dimensional bilinear game and $\s = 1$, and denote the spectral radius given $\a, \b$ as $r(\a, \b)$. If we fix $\a = 1/2$, by numerically solving \cref{ogd_j} we have
\begin{equation}
    r(1/2, 1/4) \approx 0.966,\, r(1/2, 1/3) \approx 0.956,
\end{equation}
i.e, $\a = 1/2, \b = 1/3$ is a better choice than $\a = 2\b = 1/2$.

In \Cref{schur_ana} we have seen that Jacobi momentum does not converge. Now we find the optimal convergence rate for GS momentum with a special setting of $\b_1, \b_2$, as also used in \cite{GidelHPHLLM19}.

\begin{restatable}[\textbf{momentum optimal}]{theorem}{HBopt}\label{hb_opt}
For Gauss--Seidel momentum with $\b_1 = -1/2$, $\b_2 = 0$, the optimal exponent of linear convergence is $r_*\sim 1 - 2/(9\k^2)$.
\end{restatable}

\begin{figure*}
    \centering
    \includegraphics[width=12cm]{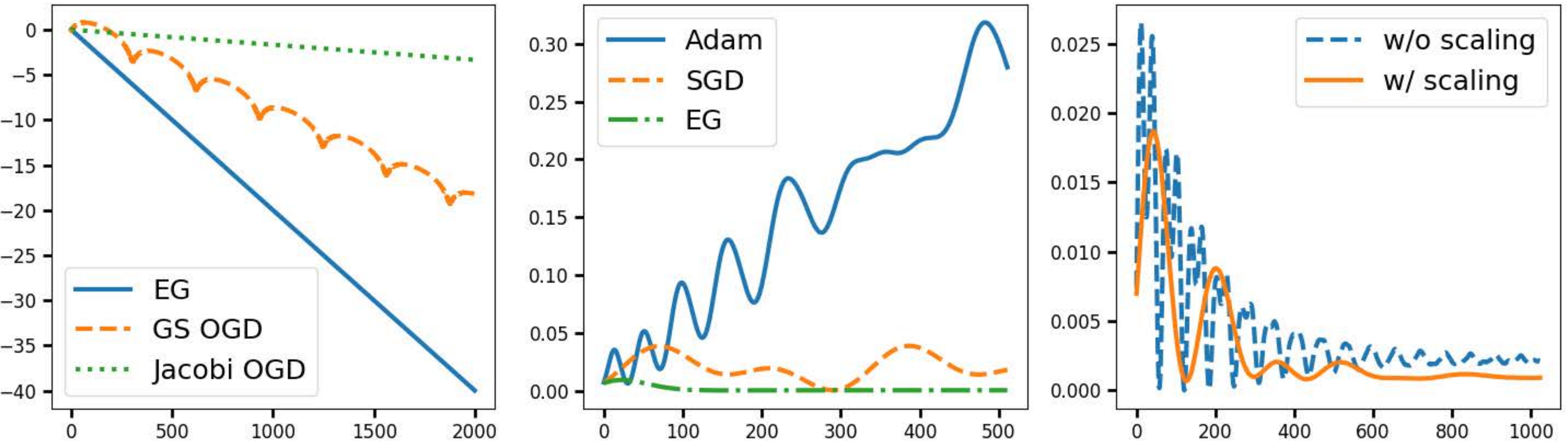}
    \caption{\textbf{Left:} linear convergence of optimal EG, Jacobi OGD, Gauss--Seidel OGD in a bilinear game with the log distance; \textbf{Middle:} comparison among Adam, SGD and EG in learning the mean of a Gaussian with WGAN with the squared distance; \textbf{Right:} Comparison between EG with ($\alpha = 0.02$, $\gamma = 2.0$) and without scaling ($\alpha = \gamma = 0.2$). We use the squared distance. }
    \label{fig:combined1}
\end{figure*}

\paragraph{Numerical method}
We provide a numerical method for finding the optimal exponent of linear convergence, by realizing that the \emph{unit} disk in Theorem \ref{schur} is not special. Let us call a polynomial to be $r$-Schur stable if all of its roots lie within an (open) disk of radius $r$ in the complex plane. We can scale the polynomial with the following lemma:
\begin{restatable}{lemma}{rSchur}\label{rSchur}
A polynomial $p(\l)$ is $r$-Schur stable iff $p(r\l)$ is Schur stable.
\end{restatable}

With the lemma above, one can rescale the Schur conditions and find the convergence region where the exponent of linear convergence is at most $r$ ($r < 1$). A simple binary search would allow one to find a better and better convergence region. See details in \Cref{appen:num}.


\section{Experiments}\label{sec:exp}
\paragraph{Bilinear game} We run experiments on a simple bilinear game and choose the optimal parameters as suggested in  \Cref{eg_opt} and \ref{ogd_opt}.
The results are shown in the left panel of \Cref{fig:combined1}, which confirms the predicted linear rates.

\begin{figure}
    \centering
    \includegraphics[width=11cm]{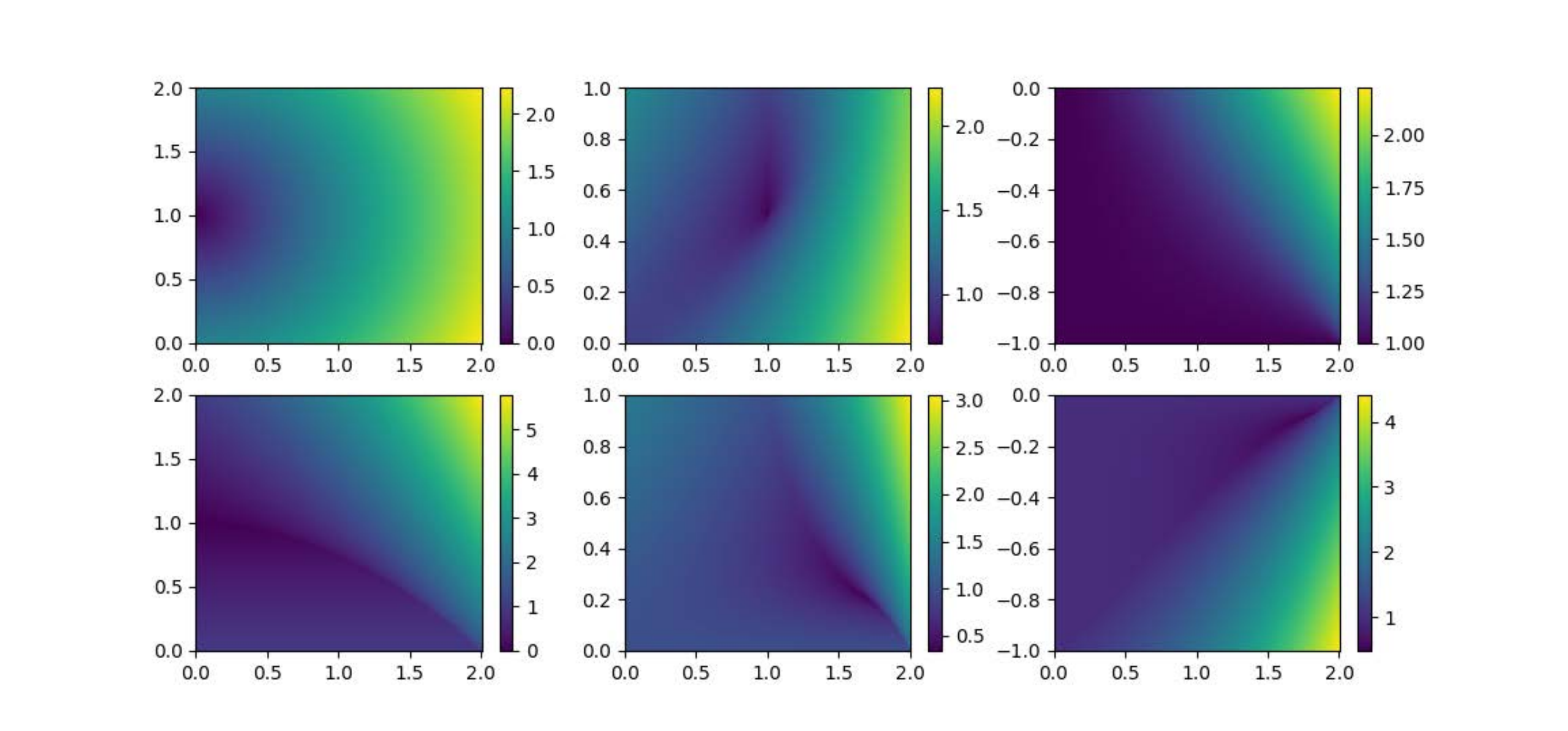}
    \caption{Heat maps of the spectral radii of different algorithms. We take $\sigma = 1$ for convenience. The \red{horizontal axis} is $\a$ and the \red{vertical axis} is $\beta$. \bb{Top row: } Jacobi updates; \bb{Bottom row: } Gauss--Seidel updates. \bb{Columns} (left to right): EG; OGD; momentum. If the spectral radius is strictly less than one, it means that our algorithm converges. In each column, the Jacobi convergence region is contained in the GS convergence region (for EG we need an additional assumption, see  \Cref{thm_eg}).}
    \label{fig:dp}
\end{figure}

\paragraph{Density plots} We show the density plots (heat maps) of the spectral radii in  \Cref{fig:dp}. We make plots for EG, OGD and momentum with both Jacobi and GS updates. These plots are made when $\b_1 = \b_2 = \b$ and they agree with our theorems in  \S\ref{schur_ana}.

\paragraph{Wasserstein GAN} As in \citet{daskalakis2017training}, we consider a WGAN \citep{arjovsky2017wasserstein} that learns the mean of a Gaussian:
\begin{align}
\textstyle
\min_{\vphi} \max_{\vtheta} f(\vphi, \vtheta) :=  \E_{\vx\sim \N(\vv, \sigma^2 \mI)} [s(\vtheta^\top \vx)] - \E_{\vz\sim \N({\bf 0}, \sigma^2 \mI)} [s(\vtheta^\top (\vz + \vphi))],
\end{align}
where $s(x)$ is the sigmoid function. It can be shown that near the saddle point $(\vtheta^*, \vphi^*) = ({\bf 0}, \vv)$ the min-max optimization can be treated as a bilinear game (\Cref{appen:shift}). With GS updates, we find that Adam diverges, SGD goes around a limit cycle, and EG converges, as shown in the middle panel of  \Cref{fig:combined1}. We can see that Adam does not behave well even in this simple task of learning a single two-dimensional Gaussian with GAN.

\begin{figure*}
    \centering
    \includegraphics[width=11cm]{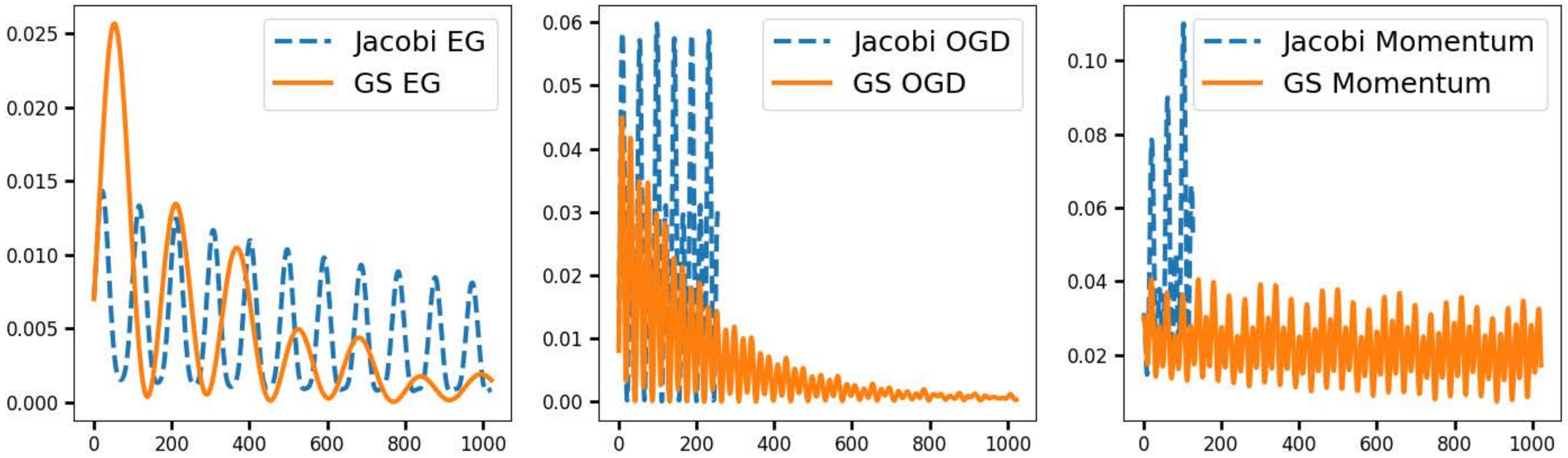}
    \caption{Jacobi vs.~GS updates. \textbf{y-axis:} Squared distance $||\vphi - \vv||^2$. \textbf{x-axis:} Number of epochs. \textbf{Left: }EG with $\gamma = 0.2, \a = 0.02$; \textbf{Middle: } OGD with $\a = 0.2$, $\beta_1 = 0.1$, $\beta_2 = 0$; \textbf{Right: }Momentum with $\a = 0.08$, $\beta = -0.1$. We plot only a few epochs for Jacobi if it does not converge.}
    \label{fig:jvsgs}
    \vspace{-1em}
\end{figure*}
Our next experiment shows that generalized algorithms may have an advantage over traditional ones. Inspired by  \Cref{eg_opt}, we compare the convergence of two EGs with the same parameter $\beta = \a \gamma$, and find that with scaling, EG has better convergence, as shown in the right panel of \Cref{fig:combined1}. Finally, we compare Jacobi updates with GS updates. In  \Cref{fig:jvsgs}, we can see that GS updates converge even if the corresponding Jacobi updates do not.

\begin{figure}
    \centering
    \includegraphics[width=10cm]{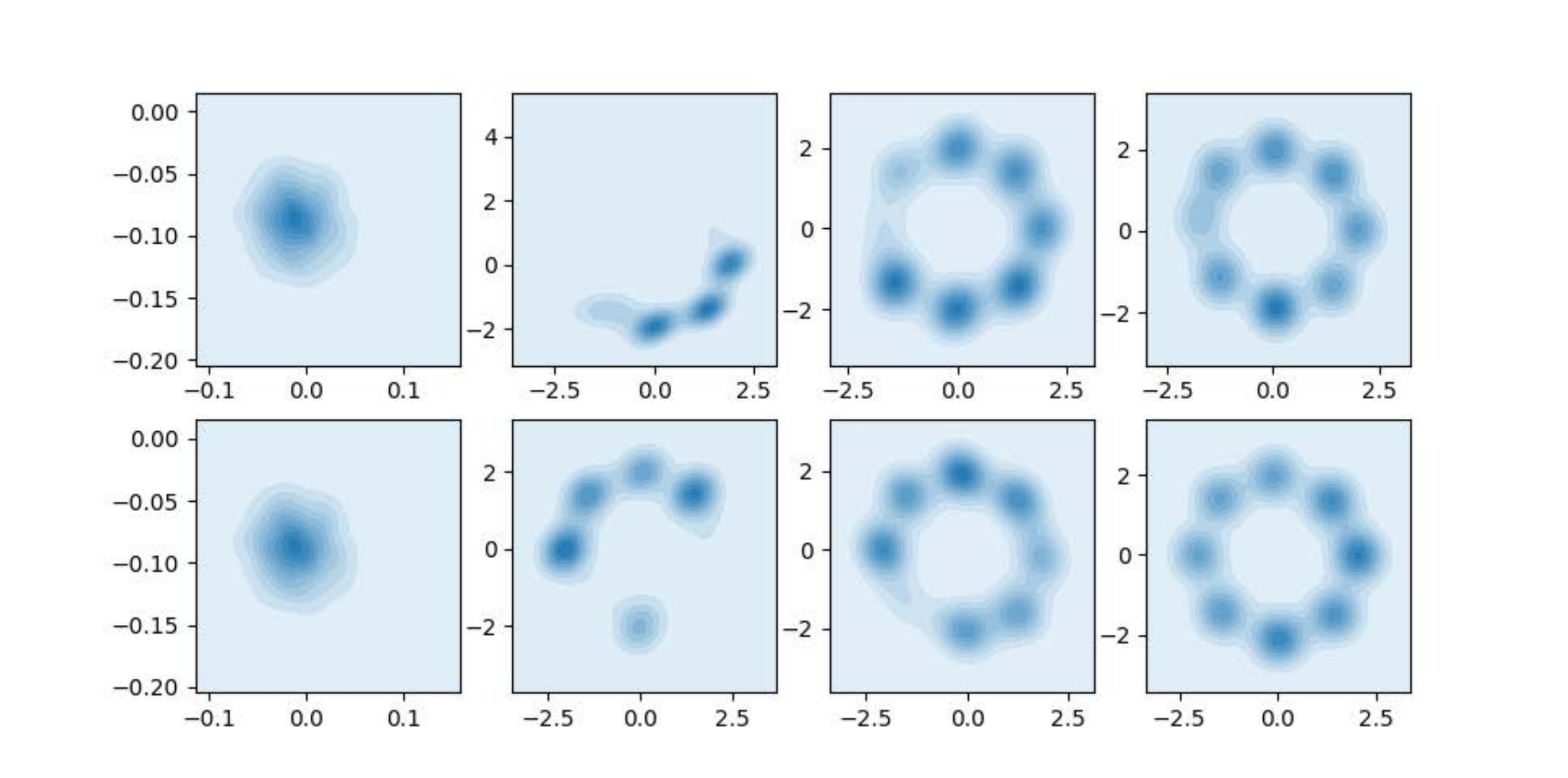}
    \caption{Test samples from the generator network trained with stochastic GD (step size $\a = 0.01$). \bb{Top row: } Jacobi updates; \bb{Bottom row: } Gauss--Seidel updates. \bb{Columns}: epoch 0, 10, 15, 20.}
    \label{fig:sgd}
\end{figure}

\begin{figure}
    \centering
    \includegraphics[width=10cm]{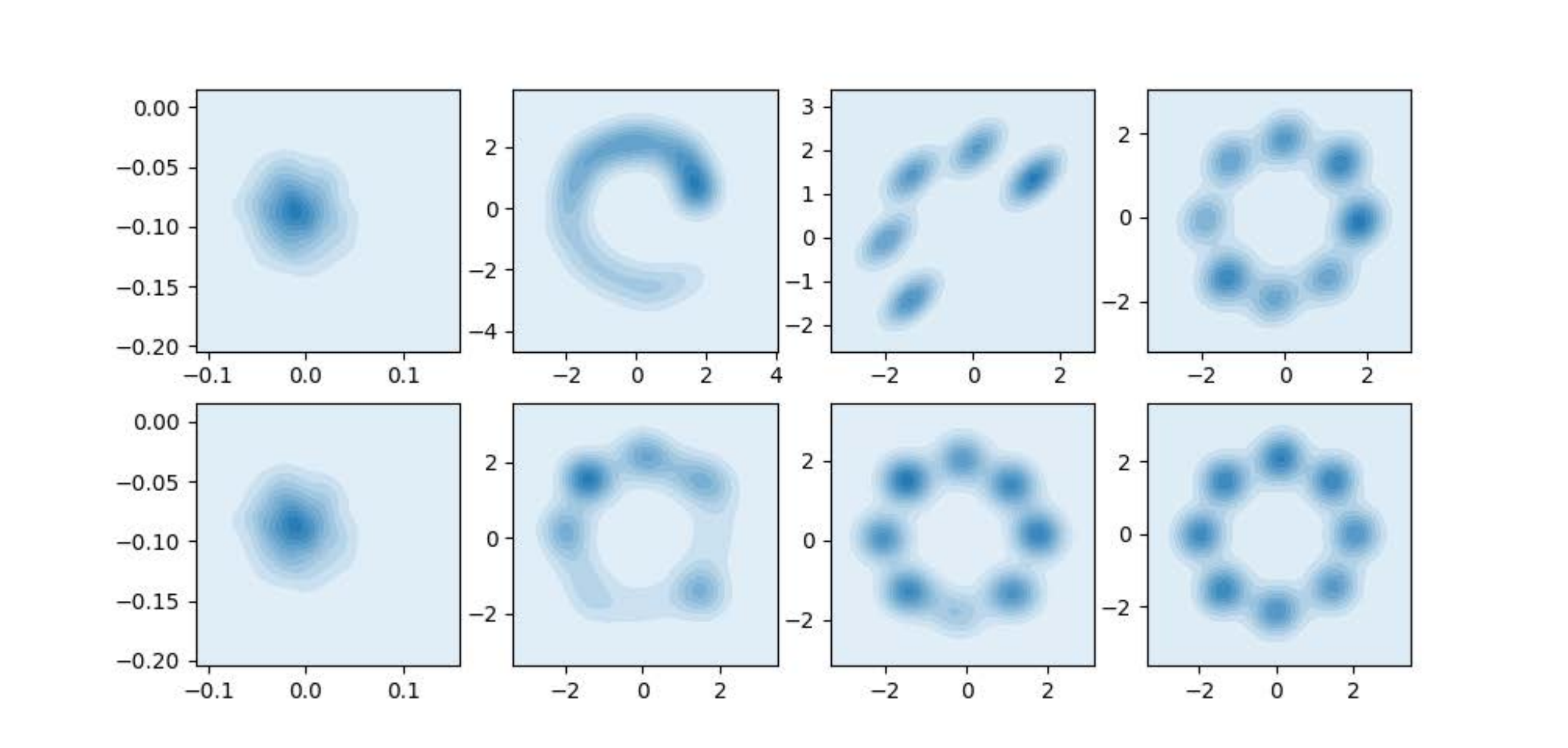}
    \caption{Test samples from the generator network trained with stochastic OGD ($\a = 2\b = 0.02$). \bb{Top row: } Jacobi updates; \bb{Bottom row: } Gauss--Seidel updates. \bb{Columns}: epoch 0, 10, 60, 100.}
    \label{fig:ogd}
\end{figure}

\paragraph{Mixtures of Gaussians (GMMs)} Our last experiment is on learning GMMs with a vanilla GAN \citep{goodfellow2014generative} that does not directly fall into our analysis. We choose a 3-hidden layer ReLU network for both the generator and the discriminator, and each hidden layer has 256 units. We find that for GD and OGD, Jacobi style updates converge more slowly than GS updates, and whenever Jacobi updates converge, the corresponding GS updates converges as well. These comparisons can be found in  \Cref{fig:sgd} and \ref{fig:ogd}, which implies the possibility of extending our results to non-bilinear games. 
Interestingly, we observe that even Jacobi GD converges on this example. We provide additional comparison between the Jacobi and GS updates of Adam \citep{kingma2014adam} in \Cref{app:adam}.

\section{Conclusions}\label{sec:conc}
In this work we focus on the convergence behaviour of gradient-based algorithms for solving bilinear games. By drawing a connection to discrete linear dynamical systems (\S \ref{prem}) and using Schur's theorem, we provide necessary and sufficient conditions for a variety of gradient algorithms, for both simultaneous (Jacobi) and alternating (Gauss--Seidel) updates. Our results show that Gauss--Seidel updates converge more easily than Jacobi updates. Furthermore, we find the optimal exponents of linear convergence for EG, OGD and the momentum method, and provide a numerical method for searching that exponent. We performed a number of experiments to validate our theoretical findings and suggest further analysis. 


There are many future directions to explore. For example, our preliminary experiments on GANs  suggest that similar (local) results might be obtained for more general games. Indeed, the local convergence behaviour of min-max nonlinear optimization can be studied through analyzing the spectrum of the Jacobian matrix of the update operator (see, e.g., \citet{nagarajan2017gradient, GidelHPHLLM19}). We believe our framework that draws the connection to linear discrete dynamic systems and Schur's theorem is a powerful machinery that can be applied in such problems and beyond.
It would be interesting to generalize our results to the constrained case (even for bilinear games), as studied in \citet{DaskalakisPanageas19, carmon2019variance}. \red{Extending our results to account for stochastic noise (as empirically tested in our experiments) is another interesting direction, with results in \citet{gidel2018variational, hsieh2019convergence}.}

\section*{Acknowledgements}
We would like to thank Argyrios Deligkas, Sarath Pattathil and Georgios Piliouras for pointing out several related references. GZ is supported by David R. Cheriton Scholarship. We gratefully acknowledge funding support from NSERC and the Waterloo-Huawei Joint Innovation Lab.

\bibliography{iclr2020_conference}
\bibliographystyle{iclr2020_conference}

\newpage
\appendix
\section{Proximal point (PP) algorithm}\label{app:pp}
PP was originally proposed by \citet{Martinet70} with $
\a_1 = \a_2$ and then carefully studied by \citet{rockafellar1976monotone}. The linear convergence for bilinear games was also proved in the same reference. Note that we do not consider Gauss--Seidel PP since we do not get a meaningful solution after a shift of steps\footnote{If one uses inverse operators this is in principle doable.}. 
\be\label{eq:pp}
\vx^{(t+1)} = \vx^{(t)} - \a_1 \gr_{\vx} f(\vx^{(t+1)}, \vy^{(t+1)}), \, \vy^{(t+1)} = \vy^{(t)} + \a_2 \gr_{\vy} f(\vx^{(t+1)}, \vy^{(t+1)}),
\en
where $\vx^{(t+1)}$ and $\vy^{(t+1)}$ are given implicitly by solving the equations above. For bilinear games, one can derive that:
\begin{eqnarray}
    && \z{t+1} = \begin{bmatrix}
     \mI & \a_1 \mE \\
     -\a_2 \mE^\top & \mI
    \end{bmatrix}^{-1} \z{t}.
\end{eqnarray}
We can compute the exact form of the inverse matrix, but perhaps an easier way is just to compute the spectrum of the original matrix (the same as Jacobi GD except that we flip the signs of $\a_i$) and perform $\l\to 1/\l$. Using the fact that the eigenvalues of a matrix are reciprocals of the eigenvalues of its inverse, the characteristic equation is:
\begin{equation}
    (1/\l - 1)^2 + \a_1 \a_2 \s^2 = 0.
\end{equation}
With the scaling symmetry $(\a_1, \a_2) \to (t \a_1, \a_2/t)$, we can take $\a_1 = \a_2 = \a > 0$. With the notations in Corollary \ref{schur_234}, we have $a = -2/(1+\a^2 \s^2)$ and $b = 1/(1+\a^2 \s^2)$, and it is easy to check $|a| < 1 + b$ and $b < 1$ are always satisfied, which means linear convergence is always guaranteed. Hence, we have the following theorem:
\begin{theorem}
For bilinear games, the proximal point algorithm always converges linearly. 
\end{theorem}

Although the proximal point algorithm behaves well, it is rarely used in practice since it is an implicit method, i.e., one needs to solve $(\x{t+1}, \y{t+1})$ from \eqref{eq:pp}. 

\section{Proofs in \Cref{prem}}

\subsection{Proof of \Cref{thm:JacobGS}}\label{pf:JacobiGS}
In this section we apply \Cref{thm:spec} to prove  \Cref{thm:JacobGS}, an interesting connection between Jacobi and Gauss--Seidel updates:
\JGS*
Let us first consider the \emph{block} linear iterative process in the sense of Jacobi (i.e., all blocks are updated \emph{simultaneously}):
\begin{align}
\label{eq:blip}
    \vz^{(t)} = \begin{bmatrix}
    \vz_1^{(t)} \\
    \vdots\\
    \vz_b^{(t)} 
    \end{bmatrix}
    = 
    \sum_{i=1}^{k} \mA_i \begin{bmatrix}
    \vz_1^{(t-i)} \\
    \vdots\\
    \vz_b^{(t-i)} 
    \end{bmatrix}
    = 
    \sum_{i=1}^{k} \left[\sum_{j=1}^{l-1} \mA_{i,j} \vz_j^{(t-i)} + \sum_{j=l}^{b} \mA_{i,j} \vz_j^{(t-i)} \right] + \vd,
\end{align}
where $\mA_{i,j}$ is the $j$-th  column block of $\mA_i$. 
For each matrix $\mA_i$, we decompose it into the sum 
\begin{align}
\mA_i = \mL_i + \mU_i,
\end{align}
where $\mL_i$ is the strictly lower \emph{block} triangular part and $\mU_i$ is the upper (including diagonal) \emph{block} triangular part.
\Cref{thm:spec} indicates that the convergence behaviour of \eqref{eq:blip} is governed by the largest modulus of the roots of the characteristic polynomial:
 \begin{align}
 \label{eq:Jacob}
\det\left( -\lambda^k \mI + \sum_{i=1}^{k} \mA_i \lambda^{k-i} \right) = \det\left(-\lambda^k \mI + \sum_{i=1}^{k} (\mL_i + \mU_i) \lambda^{k-i} \right).
 \end{align}
Alternatively,  we can also consider the updates in the sense of Gauss--Seidel (i.e., blocks are updated \emph{sequentially}):
\begin{align}
\label{eq:GS}
    \vz_l^{(t)} = \sum_{i=1}^{k} \left[\sum_{j=1}^{l-1} \mA_{i,j} \vz_j^{(t-i+1)} + \sum_{j=l}^{b} \mA_{i,j} \vz_j^{(t-i)}\right]_l + \vd_l, \quad l = 1, \ldots, b.
\end{align}
We can rewrite the Gauss--Seidel update elegantly\footnote{This is well-known when $k=1$, see e.g. \cite{Saad03}.} as:
\begin{align}
    (\mI-\mL_{1}) \vz^{(t)} = \sum_{i=1}^{k}(\mL_{i+1} + \mU_i) \vz^{(t-i)} + \vd,
\end{align}
i.e.,
\be
 \vz^{(t)} = \sum_{i=1}^{k} (\mI-\mL_{1})^{-1}(\mL_{i+1} + \mU_i) \vz^{(t-i)} + (\mI - \mL_1)^{-1}\vd, 
\en
where $\mL_{k+1} := \zero$. Applying \Cref{thm:spec} again we know the convergence behaviour of the Gauss--Seidel update is governed by the largest modulus of roots of the characteristic polynomial: 
 \begin{align}
    &\det\left( -\lambda^k \mI + \sum_{i=1}^{k} (\mI-\mL_{1})^{-1}(\mL_{i+1} + \mU_i) \lambda^{k-i} \right) \\
    &= \det\left( (\mI-\mL_{1})^{-1} \Big(-\lambda^k \mI + \lambda^k \mL_{1} + \sum_{i=1}^{k} (\mL_{i+1} + \mU_i) \lambda^{k-i} \Big)\right) \\
    &= \det(\mI-\mL_{1})^{-1} \cdot \det\left( \sum_{i=0}^{k} (\lambda \mL_i + \mU_i) \lambda^{k-i} \right) 
 \end{align}
Note that $\mA_0 = -\mI$ and the factor $\det(\mI-\mL_{1})^{-1}$ can be discarded since multiplying a characteristic polynomial by a non-zero constant factor does not change its roots.

\subsection{Proof of \Cref{schur_234}}\label{proof_corollary}
\quadpoly*

\begin{proof}
It suffices to prove the result for quartic polynomials. We write down the matrices:
\be
&&\mP_1 = [1],\, \mQ_1 = [d], \\
&&\mP_2 = \begin{bmatrix}
1 & 0 \\
a & 1
\end{bmatrix}, \,
\mQ_2 = \begin{bmatrix}
d & c \\
0 & d
\end{bmatrix},  \\
&&\mP_3 = \begin{bmatrix}
1 & 0 & 0 \\
a & 1 & 0 \\
b & a & 1
\end{bmatrix}, 
\mQ_3 = \begin{bmatrix}
d & c & b\\
0 & d & c \\
0 & 0 & d 
\end{bmatrix}, \\
&&\mP_4 = \begin{bmatrix}
1 & 0 & 0 & 0 \\
a & 1 & 0 & 0 \\
b & a & 1 & 0 \\
c & b & a & 0 
\end{bmatrix}, 
\mQ_4 = \begin{bmatrix}
d & c & b & a\\
0 & d & c & b\\
0 & 0 & d & c \\
0 & 0 & 0 & d 
\end{bmatrix}.
\en
We require $\det(\mP_k \mP_k^{\top} - \mQ_k^{\top}\mQ_k) =: \delta_k > 0$, for $k = 1, 2, 3, 4$. If $k=1$, we have $1 - d^2 > 0$, namely, $|d| < 1$. $\d_2 > 0$ reduces to $(c - a d)^2 < (1 - d^2)^2$ and thus $|c - ad| < 1 - d^2$ due to the first condition. $\d_4 > 0$ simplifies to:
\be
-((a+c)^2 - (b + d + 1)^2) ((b - d - 1) (d - 1)^2 - (a - c) (c - a d))^2 < 0, 
\en
which yields $|a + c| < |b + d + 1|$. Finally, $\d_3 > 0$ reduces to:
\be\label{appendix:schur_4}
((b - d - 1) (d - 1)^2 - (a - c) (c - a d)) ((d^2 - 1) (b + d + 1) + (c - a d) (a + c)) > 0. 
\en
Denote $p(\l) := \l^4 + a \l^3 + b \l^2 + c\l + d$, we must have $p(1) > 0$ and $p(-1) > 0$, as otherwise there is a real root $\l_0$ with $|\l_0| \geq 1$. Hence we obtain $b + d + 1 > |a + c| > 0$. Also, from $|c - ad| < 1 - d^2$, we know that:
\be
|c - ad |\cdot |a + c| <  |b +d + 1|(1-d^2) = (b + d + 1)(1-d^2).
\en
So, the second factor in \ref{appendix:schur_4} is negative and the positivity of the first factor reduces to:
\be
b < (1+d) + \frac{(c- a d)(a-c)}{(d-1)^2}.   
\en
To obtain the Schur condition for cubic polynomials, we take $d = 0$, and the quartic Schur condition becomes:
\be
|c|<1, \, |a + c| < b + 1, \, b - a c < 1 - c^2.
\en
To obtain the Schur condition for quadratic polynomials, we take $c = 0$ in the above and write:
\be
b < 1, \, |a| < 1 + b.
\en
The proof is now complete.
\end{proof}

\section{Proofs in  \Cref{schur_ana}}
Some of the following proofs in  \Cref{appen:schur_ogd} and \ref{append:pf_momen} rely on Mathematica code (mostly with the built-in function \texttt{Reduce}) but in principle the code can be verified manually using cylindrical algebraic decomposition.\footnote{See the \href{https://reference.wolfram.com/language/tutorial/SomeNotesOnInternalImplementation.html}{\blue{online Mathematica documentation}}.
}

\subsection{Derivation of characteristic polynomials}\label{appe:char_poly}
In this appendix, we derive the exact forms of LDSs (\cref{eq:LDS}) and the characteristic polynomials for all gradient-based methods introduced in \S\ref{prem}, with \cref{sec2.2: char_poly}. The following lemma is well-known and easy to verify using  Schur's complement:
\begin{lemma}
\label{block_lemma}
Given $\mM\in \R^{2n\times 2n}$, $\mA\in \R^{n\times n}$ and
\begin{equation}
    \mM = \begin{bmatrix} 
    \mA & \mB \\
    \mC & \mD 
    \end{bmatrix}.
\end{equation}
If $\mC$ and $\mD$ commute, then $\det \mM = \det(\mA \mD - \mB \mC)$.
\end{lemma}

\paragraph{Gradient descent} From \eqref{g_gd} the update equation of Jacobi GD can be derived as:
\be\label{gd_update}
    && \z{t+1} = \begin{bmatrix}
     \mI & -\a_1 \mE \\
     \a_2 \mE^\top & \mI 
    \end{bmatrix}\z{t},
\en
and with \Cref{block_lemma}, we compute the characteristic polynomial as in \cref{sec2.2: char_poly}:
\be
\det\begin{bmatrix} 
(\l - 1)\mI & \a_1 \mE \\
-\a_2 \mE^\top & (\l - 1)\mI
\end{bmatrix} = \det[(\l - 1)^2 \mI + \a_1 \a_2 \mE \mE^\top],
\en
With spectral decomposition we obtain \eqref{gd_j}. Taking $\a_2\to \l \a_2$ and with  \Cref{thm:JacobGS} we obtain the corresponding GS updates. Therefore, the characteristic polynomials for GD are:
\begin{eqnarray}
\textrm{J:}\;(\l - 1)^2 + \a_1 \a_2 \sigma^2 = 0,\, \textrm{GS:}\;(\l - 1)^2 + \a_1 \a_2 \sigma^2 \l = 0.
\end{eqnarray}

\paragraph{Extra-gradient} From \cref{g_eg_1} and \cref{g_eg_2}, the update of Jacobi EG is:
   \be
   \label{eg_j_mat} && \z{t+1} = \begin{bmatrix}
     \mI - \b_2 \mE \mE^\top & -\a_1 \mE \\
     \a_2 \mE^\top & \mI - \b_1 \mE^\top \mE 
    \end{bmatrix}\z{t},
    \en
the characteristic polynomial is:
\be
\det \begin{bmatrix} 
(\l - 1)\mI + \b_2 \mE\mE^\top & \a_1 \mE \\
-\a_2 \mE^\top & (\l - 1)\mI + \b_1 \mE^\top \mE
\end{bmatrix}.
\en
Since we assumed $\a_2 > 0$, we can left multiply the second row by $\b_2 \mE/\a_2$ and add it to the first row. Hence, we obtain:
\be
\det \begin{bmatrix} 
(\l - 1)\mI  & \a_1 \mE + (\l - 1)\b_2 \mE/\a_2 + \b_1 \b_2 \mE \mE^\top \mE/\a_2 \\
-\a_2 \mE^\top & (\l - 1)\mI + \b_1 \mE^\top \mE
\end{bmatrix}.
\en
With \Cref{block_lemma} the equation above becomes:
\be
\det [(\l-1)^2 \mI + (\b_1 + \b_2) \mE^\top \mE (\l -1) + (\a_1 \a_2 \mE^\top \mE + \b_1 \b_2 \mE^\top \mE \mE^\top \mE)],
\en
which simplifies to  \eqref{eg_j} with spectral decomposition. Note that to obtain the GS polynomial, we simply take $\a_2 \to \l \a_2$ in the Jacobi polynomial as shown in \Cref{thm:JacobGS}. For the ease of reading we copy the characteristic equations for generalized EG:
\be
   && \textrm{J:}\; (\l - 1)^2 + (\b_1 + \b_2)\sigma^2(\l - 1) + (\a_1 \a_2 \s^2 + \b_1 \b_2 \s^4) = 0,\\
   && \textrm{GS:}\; (\l - 1)^2 + (\a_1 \a_2 + \b_1 + \b_2)\sigma^2(\l - 1) + (\a_1 \a_2 \s^2 + \b_1 \b_2 \s^4) = 0.
\en

\paragraph{Optimistic gradient descent}
We can compute the LDS for OGD with \cref{g_ogd_1} and \cref{g_ogd_2}:
\be
 \label{ogd} \z{t+2} = \begin{bmatrix}
     \mI & -\a_1 \mE \\
     \a_2 \mE^\top & \mI
    \end{bmatrix}\z{t+1} +  \begin{bmatrix}
     {\bf 0} & \b_1 \mE \\
     -\b_2 \mE^\top & {\bf 0}
    \end{bmatrix}\z{t}, 
\en
With \cref{sec2.2: char_poly}, the characteristic polynomial for Jacobi OGD is
\be
\det\begin{bmatrix} 
(\l^2 - \l) \mI & (\l \a_1 - \b_1) \mE \\
(-\l \a_2 + \b_2) \mE^\top & (\l^2 - \l) \mI 
\end{bmatrix}.
\en
Taking the determinant and with \Cref{block_lemma} we obtain \eqref{ogd_j}. The characteristic polynomial for GS updates in \eqref{ogd_gs} can be subsequently derived with \Cref{thm:JacobGS}, by taking $(\a_2, \b_2) \to (\l \a_2, \l \b_2)$. For the ease of reading we copy the characteristic polynomials from the main text as:
\begin{eqnarray}
&&\textrm{J:}\;\l^2 (\l - 1)^2 + (\l \a_1 - \b_1)(\l \a_2 - \b_2)  \s^2 = 0,\\
&&\textrm{GS:}\;\l^2 (\l - 1)^2 + (\l \a_1 - \b_1)(\l \a_2 - \b_2) \l \s^2 = 0.
\end{eqnarray}

\paragraph{Momentum method} 
With \cref{g_hb_1} and \cref{g_hb_2}, the LDS for the momentum method is:
\begin{eqnarray}\label{eq:momen}
 \z{t+2} = \begin{bmatrix}
     (1+\b_1)\mI & -\a_1 \mE \\
     \a_2 \mE^\top & (1+\b_2)\mI
    \end{bmatrix}\z{t+1} +  \begin{bmatrix}
      -\b_1 \mI & {\bf 0} \\
      {\bf 0} & -\b_2 \mI 
    \end{bmatrix}\z{t}, 
\end{eqnarray}
From \cref{sec2.2: char_poly}, the characteristic polynomial for Jacobi momentum is
\be
\det\begin{bmatrix} 
(\l^2 - \l(1+\b_1) + \b_1) \mI & \l \a_1 \mE \\
-\l \a_2 \mE^\top & (\l^2 - \l(1+\b_2) + \b_2) \mI
\end{bmatrix}.
\en
Taking the determinant and with  \Cref{block_lemma} we obtain \eqref{hb_j}, while \eqref{hb_gs} can be derived with  \Cref{thm:JacobGS}, by taking $\a_2 \to \l \a_2$. For the ease of reading we copy the characteristic polynomials from the main text as:
\begin{eqnarray}
&&\textrm{J: }(\l - 1)^2 (\l - \b_1)(\l - \b_2) + \a_1 \a_2 \s^2 \l^2 = 0,\\
&&\textrm{GS: }(\l - 1)^2 (\l - \b_1)(\l - \b_2) + \a_1 \a_2 \s^2 \l^3 = 0.
\end{eqnarray}

\subsection{Proof of  \Cref{thm-gd}: Schur conditions of GD}\label{appen:schur_gd}

\GD*
\begin{proof}
With the notations in  \Cref{schur_234}, for Jacobi GD, $b = 1 + \a^2 \s^2$ > 1. For Gauss--Seidel GD, $b = 1$. The Schur conditions are violated. 
\end{proof}

\subsection{Proof of  \Cref{thm_eg}: Schur conditions of EG}\label{appen:schur_eg}

\EG*
Both characteristic polynomials can be written as a quadratic polynomial $\l^2 + a \l + b$, where:
\be
&&\textrm{J:}\; a = (\b_1 + \b_2)\s^2 - 2, \, b = (1-\b_1 \s^2)(1-\b_2 \s^2) + \a^2 \s^2,\\
&&\textrm{GS:}\; a = (\b_1 + \b_2 + \a^2)\s^2 - 2, \, b = (1-\b_1 \s^2)(1-\b_2 \s^2).
\en

Compared to Jacobi EG, the only difference between Gauss--Seidel and Jacobi updates is that the $\a^2 \s^2$ in $b$ is now in $a$, which agrees with  \Cref{thm:JacobGS}. 
Using  \Cref{schur_234}, we can derive the Schur conditions \eqref{eg_j_schur} and \eqref{eg_gs_schur}. 

More can be said if $\b_1+\b_2$ is small. For instance, if $\b_1 + \b_2 + \a^2 < 2 /\s_1^2$, then \eqref{eg_j_schur} implies \eqref{eg_gs_schur}. In this case, the first conditions of \eqref{eg_j_schur} and \eqref{eg_gs_schur} are equivalent, while the second condition of \eqref{eg_j_schur} strictly implies that of \eqref{eg_gs_schur}. Hence, the Schur region of Gauss--Seidel updates includes that of Jacobi updates. The same holds true if $\b_1 + \b_2 < \tfrac{4}{3\sigma_1^2}$.

More precisely, to show that the GS convergence region strictly contains that of the Jacobi convergence region, simply take $\b_1 = \b_2 = \b$. The Schur condition for Jacobi EG and Gauss--Seidel EG are separately:
\be
\label{s_j}&&\textrm{J:} \; \a^2 \s^2 + (\b \s^2 -1)^2 < 1, \\
\label{s_gs}&&\textrm{GS:} \; 0 < \b \s^2 < 2\textrm{ and }|\a \s| < 2 - \b \s^2.
\en
It can be shown that if $\b = \a^2/3$ and $\a\to 0$, \eqref{s_j} is always violated whereas \eqref{s_gs} is always satisfied. 

Conversely, we give an example when Jacobi EG converges while GS EG does not. Let $\beta_1 \sigma^2 = \beta_2\sigma^2 \equiv \tfrac32$, then Jacobi EG converges iff $\alpha^2 \sigma^2 < \tfrac34$ while GS EG converges iff $\alpha^2 \sigma^2 < \tfrac14$.

\subsection{Proof of \Cref{schur_ogd}: Schur conditions of OGD}\label{appen:schur_ogd}
In this subsection, we fill in the details of the proof of \Cref{schur_ogd}, by first deriving the Schur conditions of OGD, and then studying the relation between Jacobi OGD and GS OGD.

\OGD*
The Jacobi characteristic polynomial is now quartic in the form $\l^4 + a \l^3 + b \l^2 + c\l + d$, with
\begin{equation}\label{jac_ogd}
    a = -2, \, b = \a^2 \s^2 + 1, \, c = -\a(\b_1 + \b_2) \s^2, \, d = \b_1 \b_2 \s^2.
\end{equation}
Comparably, the GS polynomial \eqref{ogd_gs} can be reduced to a cubic one $\l^3 + a \l^2 + b \l + c$ with
\begin{equation}\label{coeff_gs_ogd}
    a = -2 + \a^2 \s^2, \, b = -\a(\b_1 + \b_2) \s^2 + 1, \, c =  \b_1 \b_2 \s^2.
\end{equation}

 First we derive the Schur conditions \eqref{jacobi_schur} and \eqref{gs_schur}. Note that other than Corollary \ref{schur_234}, an equivalent Schur condition can be read from \citet[Theorem 1]{cheng2007exact} as:

\renewcommand\thetheorem{\thesection.\arabic{theorem}}
\begin{theorem}[\textbf{\citet{cheng2007exact}}]\label{cheng_var}
A real quartic polynomial $\l^4 + a \l^3 + b \l^2 + c\l + d$ is Schur stable iff:
\begin{eqnarray}
  &&  |d| < 1, \, |a| < d + 3, \, |a + c| < b + d + 1, \tr
  \label{q_var} && (1 - d)^2 b  + c^2 - a (1 + d) c - (1 + d) (1 - d)^2 + a^2 d < 0.
\end{eqnarray}
\end{theorem}

With \eqref{jac_ogd} and \Cref{cheng_var}, it is straightforward to derive \eqref{jacobi_schur}. With \eqref{coeff_gs_ogd} and Corollary \ref{schur_234}, we can derive \eqref{gs_schur} without much effort.

Now, let us study the relation between the convergence region of Jacobi OGD and GS OGD, as given in \eqref{jacobi_schur} and \eqref{gs_schur}. Namely, we want to prove the last sentence of Theorem \ref{schur_ogd}. 
The outline of our proof is as follows. We first show that each region of $(\a, \b_1, \b_2)$ described in \eqref{jacobi_schur} (the Jacobi region) is contained in the region described in \eqref{gs_schur} (the GS region). Since we are only studying one singular value, we slightly abuse the notations and rewrite $\b_i \s$ as $\b_i$ ($i=1,2$) and $\a \s$ as $\a$. From \eqref{ogd_j} and \eqref{ogd_gs}, $\b_1$ and $\b_2$ can switch. WLOG, we assume $\b_1 \geq \b_2$. There are four cases to consider:
\begin{itemize}
    \item $\b_1 \geq \b_2 > 0$. The third Jacobi condition in \eqref{jacobi_schur} now is redundant, and we have $\a > \b_1$ or $\a < \b_2$ for both methods. Solving the quadratic feasibility condition for $\a$ gives:
\begin{equation}\label{vol_1}
    0 < \b_2 < 1, \, \b_2 \leq \b_1 < \frac{\b_2 + \sqrt{4 + 5\b_2^2}}{2(1+\b_2^2)}, \, \b_1 < \a < \frac{u + \sqrt{u^2 + t v}}{t},
\end{equation}
where $u = (\b_1 \b_2 + 1)(\b_1 + \b_2)$, $v = \b_1 \b_2(\b_1 \b_2 + 1)(\b_1 \b_2 - 3)$, $t = (\b_1^2 + 1)(\b_2^2 + 1)$. On the other hand, assume $\a > \b_1$, the first and third GS conditions are automatic. Solving the second gives: 
\begin{equation}\label{vol_2}
    0 < \b_2 < 1, \, \b_2 \leq \b_1 < \frac{-\b_2 + \sqrt{8 + \b_2^2}}{2}, \, \b_1 < \a < -\frac{1}{2}(\b_1 + \b_2) + \frac{1}{2}\sqrt{(\b_1 - \b_2)^2 + 16}.
\end{equation}
Define $f(\b_2) : = {-\b_2 + \sqrt{8 + \b_2^2}}/{2}$ and $g(\b_2) := ({\b_2 + \sqrt{4 + 5\b_2^2}})/({2(1+\b_2^2)})$, and one can show that 
\be\label{eq:fg}
f(\b_2) \geq g(\b_2).
\en
Furthermore, it can also be shown that given $0 < \b_2 < 1$ and $\b_2 \leq \b_1 < g(\b_2)$, we have 
\be\label{eq:uvt}
(u + \sqrt{u^2 + 4 v})/t < -(\b_1 + \b_2)/2 + (1/2)\sqrt{(\b_1 - \b_2)^2 + 16}.
\en

\item $\b_1 \geq \b_2 = 0$. The Schur condition for Jacobi and Gauss--Seidel updates reduces to:
\be
\label{b2_0_j}&&\textrm{Jacobi: }0 < \b_1 < 1, \, \b_1 < \a < \frac{2\b_1}{1+\b_1^2},\\
\label{b2_0_gs}&&\textrm{GS: }0 < \b_1 < \sqrt{2}, \, \b_1 < \a < \frac{-\b_1+\sqrt{16 + \b_1^2}}{2}.
\en
One can show that given $\b_1 \in (0, 1)$, we have $2\b_1/(1+\b_1^2) < (-\b_1 + \sqrt{16 + \b_1^2})/2$. 

\item $\b_1 \geq 0 > \b_2$. Reducing the first, second and fourth conditions of \eqref{jacobi_schur} yields:
\begin{equation}\label{vol_1_1}
    \b_2 < 0, \, 0 < \b_1 < \frac{\b_2 + \sqrt{4 + 5\b_2^2}}{2(1+\b_2^2)}, \, \b_1 < \a < \frac{u + \sqrt{u^2 + t v}}{t}.
\end{equation}
This region contains the Jacobi region. It can be similarly proved that even within this larger region, GS Schur condition \eqref{gs_schur} is always satisfied.

\item $\b_2 \leq \b_1 < 0$. We have $u < 0$, $t v < 0$ and thus $\a < (u + \sqrt{u^2 + t v})/t < 0$. This contradicts our assumption that $\a > 0$. 

\end{itemize}
Combining the four cases above, we know that the Jacobi region is contained in the GS region. 

To show the strict inclusion, take $\b_1 = \b_2 = \a/5$ and $\a\to 0$. One can show that as long as $\a$ is small enough, all the Jacobi regions do not contain this point, each of which is described with a singular value in \eqref{jacobi_schur}. However, all the GS regions described in \eqref{gs_schur} contain this point. 

The proof above is still missing some details. We provide the proofs of \eqref{vol_1}, \eqref{eq:fg}, \eqref{eq:uvt} and \eqref{vol_1_1} in the sub-sub-sections below, with the help of Mathematica, although one can also verify these claims manually. Moreover, a one line proof of the inclusion can be given with Mathematica code, as shown in Section \ref{1l}.

\subsubsection{Proof of equation \ref{vol_1}} 
 The fourth condition of \eqref{jacobi_schur} can be rewritten as:
\be\label{eq:append:1}
\a^2 t - 2 u \a - v < 0,
\en
where $u = (\b_1 \b_2 + 1)(\b_1 + \b_2)$, $v = \b_1 \b_2(\b_1 \b_2 + 1)(\b_1 \b_2 - 3)$, $t = (\b_1^2 + 1)(\b_2^2 + 1)$. The discriminant is $4(u^2 + t v) = (1 - \b_1 \b_2)^2 (1 + \b_1 \b_2)(\b_1^2 + \b_2^2 + \b_1^2 \b_2^2 - \b_1 \b_2) \geq 0$. Since if $\b_1 \b_2 < 0$, 
$$
\b_1^2 + \b_2^2 + \b_1^2 \b_2^2 - \b_1 \b_2 = \b_1^2 + \b_2^2 + \b_1 \b_2 (\b_1 \b_2 - 1) > 0, 
$$
If $\b_1 \b_2 \geq 0$, 
$$
\b_1^2 + \b_2^2 + \b_1^2 \b_2^2 - \b_1 \b_2 =(\b_1 - \b_2)^2 + \b_1 \b_2 (1 + \b_1 \b_2) \geq 0,
$$
where we used $|\b_1 \b_2| < 1$ in both cases. So, \eqref{eq:append:1} becomes:
\be\label{eq:4_j_ogd}
\frac{u - \sqrt{u^2 + t v}}{t} < \a <  \frac{u + \sqrt{u^2 + t v}}{t}.
\en
Combining with $\a > \b_1$ or $\a < \b_2$ obtained from the second condition, we have:
\be
\frac{u - \sqrt{u^2 + t v}}{t} < \a  < \b_2\textrm{ or } \b_1 < \a <  \frac{u + \sqrt{u^2 + t v}}{t}.
\en
The first case is not possible, with the following code:
\begin{verbatim}
    u = (b1 b2 + 1) (b1 + b2); v = b1 b2 (b1 b2 + 1) (b1 b2 - 3);
    t = (b1^2 + 1) (b2^2  + 1);
    Reduce[b2 t > u - Sqrt[u^2 + t v] && b1 >= b2 > 0 
    && Abs[b1 b2] < 1],
\end{verbatim}
and we have:
\begin{verbatim}
    False.
\end{verbatim}
Therefore, the only possible case is $\b_1 < \a <  ({u + \sqrt{u^2 + t v}})/{t}$. Where the feasibility region can be solved with:
\begin{verbatim}
    Reduce[b1 t < u + Sqrt[u^2+t v]&&b1>=b2>0&&Abs[b1 b2] < 1].
\end{verbatim}
What we get is:
\begin{verbatim}
    0<b2<1 &&
    b2<=b1<b2/(2 (1+b2^2))+1/2 Sqrt[(4+5 b2^2)/(1+b2^2)^2].
\end{verbatim}
Therefore, we have proved \eqref{vol_1}.

\subsubsection{Proof of equation \ref{eq:fg}}
With
\begin{verbatim}
    Reduce[-(b2/2) + Sqrt[8 + b2^2]/2 >= 
    (b2 + Sqrt[4 + 5 b2^2])/(2 (1 + b2^2)) && 0 < b2 < 1],
\end{verbatim}
we can remove the first constraint and get:
\begin{verbatim}
    0 < b2 < 1.
\end{verbatim}

\subsubsection{Proof of equation \ref{eq:uvt}}
Given
\begin{verbatim}
    Reduce[-1/2 (b1 + b2) + 1/2 Sqrt[(b1 - b2)^2 + 16] > 
    (u + Sqrt[u^2 + t v])/t && 
  0 < b2 < 1 && 
  b2 <= b1 < (b2 + Sqrt[4 + 5 b2^2])/(2 (1 + b2^2)), {b2, b1}],
\end{verbatim}
we can remove the first constraint and get:
\begin{verbatim}
    0 < b2 < 1 && 
 b2 <= b1 < b2/(2 (1 + b2^2)) + 
 1/2 Sqrt[(4 + 5 b2^2)/(1 + b2^2)^2].
\end{verbatim}

\subsubsection{Proof of equation \ref{vol_1_1}}
The second Jacobi condition simplifies to $\a > \b_1$ and the fourth simplifies to \eqref{eq:4_j_ogd}. Combining with the first Jacobi condition:
\begin{verbatim}
    Reduce[Abs[b1 b2] < 1 && 
   a > b1 && (u - Sqrt[u^2 + t v])/t < a < (u + Sqrt[u^2 + t v])/t 
   && b1 >= 0 && b2 < 0, {b2, b1, a} ] // Simplify,
\end{verbatim}
we have:
\begin{verbatim}
    b2 < 0 && b1 > 0 && 
 b2/(1 + b2^2) + Sqrt[(4 + 5 b2^2)/(1 + b2^2)^2] > 2 b1 && 
 b1 < a < (b1 + b2 + b1^2 b2 + b1 b2^2)/((1 + b1^2) (1 + b2^2)) + 
   Sqrt[((-1 + b1 b2)^2 (b1^2 + b2^2 + b1 b2 (-1 + b2^2) + 
      b1^3 (b2 + b2^3)))/((1 + b1^2)^2 (1 + b2^2)^2)].
\end{verbatim}
This can be further simplified to achieve \eqref{vol_1_1}.

\subsubsection{One line proof}\label{1l}
In fact, there is another very simple proof:
\begin{verbatim}
    Reduce[ForAll[{b1, b2, a}, (a - b1) (a - b2) > 0
    && (a + b1) (a + b2) > -4 && Abs[b1 b2] < 1 &&
    a^2 (b1^2 + 1) (b2^2 + 1) < (b1 b2 + 1) (2 a (b1 + b2) + 
    b1 b2 (b1 b2 - 3)), (a - b1) (a - b2) > 0 && 
    (a + b1) (a + b2) < 4
    && (a b1 + 1) (a b2 + 1) > (1 + b1 b2)^2], {b2, b1, a}]
    True.
\end{verbatim}
However, this proof does not tell us much information about the range of our variables.

\subsection{proof of \Cref{thm_momentum}: Schur conditions of momentum}\label{append:pf_momen}

\Momentum*
\subsubsection{Schur conditions of Jacobi and GS updates}
\paragraph{Jacobi condition} 
We first rename $\a \s$ as \texttt{al} and $\b_1, \b_2$ as \texttt{b1}, \texttt{b2}. With Theorem \ref{cheng_var}:
\begin{verbatim}
    {Abs[d] < 1, Abs[a] < d + 3, 
   a + b + c + d + 1 > 0, -a + b - c + d + 1 > 
    0, (1 - d)^2 b  - (c - a d) (a - c) - (1 + d) (1 - d)^2 < 
    0} /. {a -> -2 - b1 - b2, b -> al^2 + 1 + 2 (b1 + b2) + b1 b2, 
   c -> -b1 - b2 - 2 b1 b2, d -> b1 b2} // FullSimplify.
\end{verbatim}
We obtain:
\begin{verbatim}
    {Abs[b1 b2] < 1, Abs[2 + b1 + b2] < 3 + b1 b2, al^2 > 0, 
 al^2 + 4 (1 + b1) (1 + b2) > 0, al^2 (-1 + b1 b2)^2 < 0}.
\end{verbatim}
The last condition is never satisfied and thus Jacobi momentum never converges. 
\paragraph{Gauss--Seidel condition} With Theorem \ref{cheng_var}, we compute:
\begin{verbatim}
{Abs[d] < 1, Abs[a] < d + 3, 
   a + b + c + d + 1 > 0, -a + b - c + d + 1 > 
    0, (1 - d)^2 b  + c^2 - a (1 + d) c - (1 + d) (1 - d)^2 + a^2 d < 
    0} /. {a -> al^2 - 2 - b1 - b2, b -> 1 + 2 (b1 + b2) + b1 b2, 
   c -> -b1 - b2 - 2 b1 b2, d -> b1 b2} // FullSimplify.
\end{verbatim}
The result is:
\begin{verbatim}
{Abs[b1 b2] < 1, Abs[2 - al^2 + b1 + b2] < 3 + b1 b2, al^2 > 0, 
 4 (1 + b1) (1 + b2) > al^2, 
 al^2 (b1 + b2 + (-2 + al^2 - b1) b1 b2 + b1 (-1 + 2 b1) b2^2) < 0},
\end{verbatim}
which can be further simplified to \eqref{gs_momentum}.

\subsubsection{Negative momentum}

With Theorem \ref{thm_momentum}, we can actually show that in general at least one of $\b_1$ and $\b_2$ must be negative. There are three cases to consider, and in each case we simplify \eqref{gs_momentum}:
\begin{enumerate}
    \item $\b_1 \b_2 = 0$. WLOG, let $\b_2 = 0$, and we obtain 
    \be\label{b2_0_momentum}
    -1 < \b_1 < 0\textrm{ and }\a^2 \s^2 < 4(1+\b_1).
    \en
    \item $\b_1 \b_2 > 0$. We have
    \be\label{eq:momen_po}
    -1 < \b_1 < 0, \, -1 < \b_2 < 0\, , \a^2 \s^2 < 4(1+\b_1)(1+\b_2).
    \en
    \item $\b_1 \b_2 < 0$. WLOG, we assume $\b_1 \geq \b_2$. We obtain:
    \be\label{eq:mo_neg1}
    -1 < \b_2 < 0, 0 < \b_1 < \min \left\{-\frac{1}{3\b_2}, \Big\rvert-\frac{\b_2}{1+2\b_2}\Big\rvert\right\}.
    \en
    The constraints for $\a$ are $\a > 0$ and:
    \be\label{eq:mo_neg2}
    \max\left\{\frac{(1-\b_1 \b_2)(2\b_1 \b_2 - \b_1 -\b_2)} {\b_1 \b_2}, 0\right\} < \a^2 \s^2 < 4(1+\b_1)(1+\b_2).
    \en
\end{enumerate}
These conditions can be further simplified by analyzing all singular values. They only depend on $\s_1$ and $\s_n$, the largest and the smallest singular values. Now, let us derive \eqref{eq:momen_po}, \eqref{eq:mo_neg1} and \eqref{eq:mo_neg2} more carefully. Note that we use \texttt{a} for $\a \s$. 

\subsubsection{Proof of equation \ref{eq:momen_po}}

\begin{verbatim}
    Reduce[Abs[b1 b2] < 1 && Abs[-a^2 + b1 + b2 + 2] < b1 b2 + 3 && 
  4 (b1 + 1) (b2 + 1) > a^2 && 
  a^2 b1 b2 < (1 - b1 b2) (2 b1 b2 - b1 - b2) && b1 b2 > 0 && 
  a > 0, {b2, b1, a}]
\end{verbatim}

\begin{verbatim}
    -1 < b2 < 0 && -1 < b1 < 0 && 0 < a < Sqrt[4 + 4 b1 + 4 b2 + 4 b1 b2]
\end{verbatim}

\subsubsection{Proof of equations \ref{eq:mo_neg1} and \ref{eq:mo_neg2}}

\begin{verbatim}
    Reduce[Abs[b1 b2] < 1 && Abs[-a^2 + b1 + b2 + 2] < b1 b2 + 3 && 
  4 (b1 + 1) (b2 + 1) > a^2 && 
  a^2 b1 b2 < (1 - b1 b2) (2 b1 b2 - b1 - b2) && b1 b2 < 0 && 
  b1 >= b2 && a > 0, {b2, b1, a}]
\end{verbatim}

\begin{verbatim}
    (-1 < b2 <= -(1/3) && ((0 < b1 <= b2/(-1 + 2 b2) && 
       0 < a < Sqrt[4 + 4 b1 + 4 b2 + 4 b1 b2]) || (b2/(-1 + 2 b2) < 
        b1 < -(1/(3 b2)) && 
       Sqrt[(-b1 - b2 + 2 b1 b2 + b1^2 b2 + b1 b2^2 - 2 b1^2 b2^2)/(
        b1 b2)] < a < Sqrt[4 + 4 b1 + 4 b2 + 4 b1 b2]))) || (-(1/3) < 
    b2 < 0 && ((0 < b1 <= b2/(-1 + 2 b2) && 
       0 < a < Sqrt[4 + 4 b1 + 4 b2 + 4 b1 b2]) || (b2/(-1 + 2 b2) < 
        b1 < -(b2/(1 + 2 b2)) && 
       Sqrt[(-b1 - b2 + 2 b1 b2 + b1^2 b2 + b1 b2^2 - 2 b1^2 b2^2)/(
        b1 b2)] < a < Sqrt[4 + 4 b1 + 4 b2 + 4 b1 b2])))
\end{verbatim}
Some further simplication yields \eqref{eq:mo_neg1} and \eqref{eq:mo_neg2}.

\section{Proofs in Section \ref{opt_spe}}
For bilinear games and gradient-based methods, a Schur condition defines the region of convergence in the parameter space, as we have seen in Section \ref{schur_ana}. However, it is unknown which setting of parameters has the best convergence rate in a Schur stable region. We explore this problem now. Due to Theorem \ref{thm-gd}, we do not need to study GD. The remaining cases are EG, OGD and GS momentum (Jacobi momentum does not converge due to Theorem \ref{thm_momentum}). Analytically (Section \ref{app:eg} and \ref{app:ogd}), we study the optimal linear rates for EG and special cases of generalized OGD (Jacobi OGD with $\b_1 = \b_2$ and Gauss--Seidel OGD with $\b_2 = 0$). The special cases include the original form of OGD. We also provide details for the numerical method described at the end of Section \ref{opt_spe}.

The optimal spectral radius is obtained by solving another min-max optimization problem:
\be\label{small_min_max}
\min_{\vtheta} \max_{\s \in {\rm Sv}(\mE)} r(\vtheta, \s),
\en
where $\vtheta$ denotes the collection of all hyper-parameters, and $r(\vtheta, \s)$ is defined as the spectral radius function that relies on the choice of parameters and the singular value $\s$. We also use ${\rm Sv}(\mE)$ to denote the set of singular values of $\mE$.

In general, the function $r(\vtheta, \sigma)$ is non-convex and thus difficult to analyze. However, in the special case of quadratic characteristic polynomials, it is possible to solve \eqref{small_min_max}. This is how we will analyze EG and special cases of OGD, as $r(\vtheta, \s)$ can be expressed using root functions of quadratic polynomials. For cubic and quartic polynomials, it is in principle also doable as we have analytic formulas for the roots. However, these formulas are extremely complicated and difficult to optimize and we leave it for future work. For EG and OGD, we will show that the optimal linear rates depend only on the conditional number $\kappa := \s_1/\s_n$. 

For simplicity, we always fix $\a_1 = \a_2 = \a > 0$ using the scaling symmetry studied in Section \ref{schur_ana}. 

\subsection{Proof of Theorem \ref{eg_opt}: Optimal convergence rate of EG}\label{app:eg}

\renewcommand{\thetheorem}{4.1}
\begin{theorem}[\textbf{EG optimal}]
Both Jacobi and GS EG achieve the optimal exponent of linear convergence $r_* = (\k^2 - 1)/(\k^2 + 1)$ at $\a \to 0$ and $\b_1 = \b_2 = 2/(\s_1^2 + \s_n^2)$. As $\kappa \to \infty$, $r_* \to 1 - 2/\k^2$. 
\end{theorem}

\subsubsection{Jacobi EG}

For Jacobi updates, if $\b_1 = \b_2 = \b$, by solving the roots of \eqref{eg_j}, the min-max problem is:
\begin{equation}\label{jac_eg_opt}
    \min_{\a, \b} \max_{\s \in {\rm Sv}(\mE)} \sqrt{\a^2 \s^2 + (1-\b \s^2)^2}.
\end{equation}
If $\s_1 = \s_n = \s$, we can simply take $\a\to 0$ and $\b = 1/\s^2$ to obtain a super-linear convergence rate. Otherwise, let us assume $\s_1 > \s_n$. We obtain a lower bound by taking $\a \to 0$ and \eqref{jac_eg_opt} reduces to:
\be\label{jac_eg_opt_l}
 \min_{\b} \max_{\s \in {\rm Sv}(\mE)} |1 - \b \s^2|.
\en
The optimal solution is given at $1 - \b \s_n^2 = \b \s_1^2 - 1$, yielding $\b = 2/(\s_1^2 + \s_n^2)$. The optimal radius is thus $(\s_1^2 - \s_n^2)/(\s_1^2 + \s_n^2)$ since the lower bound \eqref{jac_eg_opt_l} can be achieved by taking $\a \to 0$.

From general $\b_1,\, \b_2$, it can be verified that the optimal radius is achieved at $\b_1 = \b_2$ and the problem reduces to the previous case. The optimization problem is:
\begin{equation}\label{gs_op_gen}
    \min_{\a, \b_1, \b_2} \max_{\s \in {\rm Sv}(\mE)} r(\a, \b_1, \b_2, \s),
\end{equation}
where 
$$
r(\a, \b_1, \b_2, \s) = \begin{cases}
\sqrt{(1 -\b_1 \s^2)(1-\b_2 \s^2) + \a^2 \s^2} & 4\a^2 > (\b_1-\b_2 )^2 \s^2,\\
|1-\frac{1}{2}(\b_1 + \b_2)\s^2| + \frac{1}{2}\sqrt{(\b_1 - \b_2)^2 \s^4 - 4 \a^2 \s^2}  & 4\a^2 \leq (\b_1-\b_2 )^2 \s^2.
\end{cases}
$$
In the first case, a lower bound is obtained at $\a^2 = (\b_1 - \b_2)^2 \s^2/4$ and thus the objective only depends on $\b_1 + \b_2$. In the second case, the lower bound is obtained at $\a \to 0$ and $\b_1 \to \b_2$. Therefore, the function is optimized at $\b_1 = \b_2$ and $\a \to 0$.

Our analysis above does not mean that $\a \to 0$ and $\b_1 = \b_2 = 2/(\s_1^2 + \s_n^2)$ is the only optimal choice. For example, when $\s_1 = \s_n = 1$, we can take $\b_1 = 1 + \a$ and $\b_2 = 1 - \a$ to obtain a super-linear convergence rate. 

\subsubsection{Gauss--Seidel EG}
For Gauss--Seidel updates and $\b_1 = \b_2 = \b$, we do the following optimization:
\begin{equation}\label{gs_eg_op}
    \min_{\a, \b} \max_{\s \in {\rm Sv}(\mE)} r(\a, \b, \s),
\end{equation}
where by solving \eqref{eg_gs}:
$$
r(\a, \b, \s) = \begin{cases}
1 -\b \s^2 & \a^2 \s^2 < 4(1-\b \s^2),\\
\frac{\a^2}{2}\s^2 - (1 - \b \s^2) + \sqrt{\a^2 \s^2(\a^2 \s^2 - 4(1-\b \s^2))}/2 &  \a^2 \s^2 \geq 4(1-\b \s^2).
\end{cases}
$$
$r(\s, \b, \s^2)$ is quasi-convex in $\s^2$, so we just need to minimize over $\a, \b$ at both end points. Hence, \eqref{gs_eg_op} reduces to:
$$
    \min_{\a, \b} \max \{r(\a, \b, \s_1), r(\a, \b, \s_n)\}.
$$
By arguing over three cases: $\a^2 + 4\b < 4/\s_1^2$, $\a^2 + 4\b > 4/\s_n^2$ and $4/\s_1^2 \leq \a^2 + 4\b \leq 4/\s_n^2$, we find that the minimum $(\k^2 - 1)/(\k^2 + 1)$ can be achieved at $\a\to 0$ and $\b = 2/(\s_1^2 + \s_n^2)$, the same as Jacobi EG. This is because $\a \to 0$ decouples $x$ and $y$ and it does not matter whether the update is Jacobi or GS.

For general $\b_1,\, \b_2$, it can be verified that the optimal radius is achieved at $\b_1 = \b_2$. We do the following transformation: $\b_i \to \xi_i - \a^2/2$, so that the characteristic polynomial becomes:
\begin{equation}
    (\l - 1)^2 + (\xi_1 + \xi_2)\s^2(\l - 1) + \a^2 \s^2 + (\xi_1 - \a^2/2)(\xi_2 - \a^2 /2) \s^4 = 0.
\end{equation}
Denote $\xi_1 + \xi_2 = \phi$, and $ (\xi_1 - \a^2/2)(\xi_2 - \a^2 /2) = \nu$, we have:
\begin{equation}
    \l^2- (2 - \s^2 \phi)\l + 1 - \s^2 \phi + \s^4 v + \s^2 \a^2 = 0.
\end{equation}
The discriminant is $\Delta := \s^2(\s^2(\phi^2 - 4\nu) - 4\a^2)$. We discuss two cases:
\begin{enumerate}
    \item $\phi^2 - 4\nu < 0$. We are minimizing:
    $$
    \min_{\a, u, v} \sqrt{1 + (\a^2 - \phi)\s_1^2 + \s_1^4 \nu}  \vee
    \sqrt{1 + (\a^2 - \phi)\s_n^2 + \s_n^4 \nu},
    $$
    with $a \vee b : = \max\{a, b\}$ a shorthand. A minimizer is at $\a\to 0$ and $\nu\to \phi^2/4$ (since $\phi^2 < 4\nu$), where $\b_1 = \b_2 = 2/(\s_1^2 + \s_n^2)$ and $\a \to 0$. 
    \item $\phi^2 - 4\nu \geq 0$. A lower bound is:
    $$
    \min_u |1-\phi\s_1^2/2| \vee |1-\phi\s_n^2/2|,
    $$
    which is obtained iff $4\a^2 \sim (\phi^2 - 4\nu)t$ for all $\s^2$. This is only possible if $\a\to 0$ and $\phi^2 \to 4\nu$, which yields $\b_1 = \b_2 = 2/(\s_1^2 + \s_n^2)$.
\end{enumerate}
From what has been discussed, the optimal radius is $(\k^2 - 1)/(\k^2 + 1)$ which can be achieved at $\b_1 = \b_2 = 2/(\s_1^2 + \s_n^2)$ and $\a \to 0$. Again, this might not be the only choice. For instance, take $\s_1 = \s_n^2 = 1$, from \eqref{eg_gs}, a super-linear convergence rate can be achieved at $\b_1 = 1$ and $\b_2 = 1 - \a^2$.

\subsection{Proof of Theorem \ref{ogd_opt}: Optimal convergence rate of OGD}\label{app:ogd}
\renewcommand{\thetheorem}{4.2}
\begin{theorem}[\textbf{OGD optimal}]
For Jacobi OGD with $\b_1 = \b_2 = \b$, to achieve the optimal linear rate, we must have $\a \leq 2\b$.  For the original OGD with $\a = 2\b$, the optimal linear rate $r_*$ satisfies 
\be
r_*^2 = \frac{1}{2} + \frac{1}{4\sqrt{2}\s_1^2}\sqrt{(\s_1^2 - \s_n^2)(5 \s_1^2 - \s_n^2 + \sqrt{(\s_1^2 - \s_n^2)(9 \s_1^2 - \s_n^2)})},
\en
at     
\begin{equation}
\b_* = \frac{1}{4\sqrt{2}}\sqrt{\frac{3 \s_1^4 - (\s_1^2 - \s_n^2)^{3/2}\sqrt{9\s_1^2 - \s_n^2} + 6 \s_1^2 \s_n^2 - \s_n^4}{\s_1^4 \s_n^2}}.
\end{equation}
If $\k \to \infty$, $r_* \sim 1 - 1/(6\k^2)$. For Gauss--Seidel OGD with $\b_2 = 0$, the optimal linear rate is $r_* = \sqrt{(\k^2 - 1)/(\k^2 + 1)}$, at $\a = \sqrt{2}/\s_1$ and $\b_1 = \sqrt{2}\s_1/(\s_1^2 + \s_n^2)$. If $\k \to \infty$, $r_* \sim 1 - 1/\k^2$. 
\end{theorem}

For OGD, the characteristic polynomials \eqref{ogd_j} and \eqref{ogd_gs} are quartic and cubic separately, and thus optimizing the spectral radii for generalized OGD is difficult. However, we can study two special cases: for Jacobi OGD, we take $\b_1 = \b_2$; for Gauss--Seidel OGD, we take $\b_2 = 0$. In both cases, the spectral radius functions can be obtained by solving quadratic polynomials.

\subsubsection{Jacobi OGD} 
We assume $\b_1 = \b_2 = \b$ in this subsection. The characteristic polynomial for Jacobi OGD \eqref{ogd_j} can be written as:
\be
\l^2 (\l - 1)^2 + (\l \a - \b)^2 \s^2 = 0.
\en
Factorizing it gives two equations which are conjugate to each other:
\be
\l(\l - 1) \pm i (\l \a - \b)\s = 0.
\en
The roots of one equation are the conjugates of the other equation. WLOG, we solve $\l(\l - 1) + i (\l \a - \b)\s = 0$ which gives $({1}/{2})(u \pm v)$, where
\begin{align}
u = 1 - i \a \s, \, v = \sqrt{1-\alpha^2 \s^2 - 2i(\alpha-2\beta)\s}.
\end{align}

Denote $\Delta_1 = 1 - \alpha^2 \s^2 $ and $\Delta_2 = 2(\alpha - 2\beta)\s$. If $\a \geq 2\b$, $v$ can be expressed as:
\begin{align}
v = \frac{1}{\sqrt{2}}\left( \sqrt{\sqrt{\Delta_1^2 + \Delta_2^2} + \Delta_1} - i\sqrt{\sqrt{\Delta_1^2 + \Delta_2^2} - \Delta_1} \right)  =: \frac{1}{\sqrt{2}}(a - i b),
\end{align}
therefore, the spectral radius $r(\a, \b, \s)$ satisfies:
\begin{align}
r(\a, \b, \s)^2 = \frac{1}{4}\left((1 + a/\sqrt{2})^2 + (\alpha \s + b/\sqrt{2})^2\right) = \frac{1}{4}(1 + \a^2 \s^2 + \sqrt{\Delta_1^2 + \Delta_2^2} + \sqrt{2} (b \s \a + a)),
\end{align}
and the minimum is achieved at $\alpha = 2 \beta$. From now on, we assume $\a \leq 2\b$,  and thus $v = a + i b$. We write:
\be
r(\a, \b, \s)^2 &=& \frac{1}{4}\max \{\left((1 + a/\sqrt{2})^2 + (\alpha \s - b/\sqrt{2})^2\right),
    \left((1 - a/\sqrt{2})^2 + (\alpha \s + b/\sqrt{2})^2 \right) \},\tr
&=& \frac{1}{4}(1 + \a^2 \s^2 + \sqrt{\Delta_1^2 + \Delta_2^2} + \sqrt{2}  |b \s \a - a|).\tr
&=&\begin{cases}
\frac{1}{4}(1 + \a^2 \s^2 + \sqrt{\Delta_1^2 + \Delta_2^2} - \sqrt{2}  (b \s \a - a)) & 0 < \a \s \leq 1,\\
\frac{1}{4}(1 + \a^2 \s^2 + \sqrt{\Delta_1^2 + \Delta_2^2} + \sqrt{2}  (b \s \a - a)) & \a \s > 1.
\end{cases}
\en
This is a non-convex and non-differentiable function, which is extremely difficult to optimize. 

At $\a = 2\b$, in this case, $a = \sqrt{1-4\b^2 \s^2}{\rm sign}(1-4\b^2 \s^2)$ and $b = \sqrt{4\b^2 \s^2 - 1}{\rm sign}(4\b^2 \s^2 - 1)$. The sign function ${\rm sign}(x)$ is defined to be $1$ if $x > 0$ and $0$ otherwise. The function we are optimizing is a quasi-convex function:
\begin{equation}
    r(\b, \s)^2 = \begin{cases}
    \frac{1}{2}(1 + \sqrt{1-4\b^2 \s^2}) & 4\b^2 \s^2 \leq 1,\\
    2\b^2 \s^2 + \b \s \sqrt{4\b^2 \s^2 - 1} & 4\b^2 \s^2 > 1.
    \end{cases}
\end{equation}
We are maximizing over $\s$ and minimizing over $\b$. There are three cases: 
\begin{itemize}
    \item $4\b^2 \s_1^2 \leq 1$. At $4\b^2 \s_1^2 = 1$, the optimal radius is:
    $$
    r_*^2 = \frac{1}{2}\left( 1 + \sqrt{1 - \frac{1}{\k^2}}\right).
    $$
    \item $4\b^2 \s_n^2 \geq 1$. At $4\b^2 \s_n^2 = 1$, the optimal radius satisfies: 
    $$
    r_*^2 = \frac{\k^2}{2} + \frac{\k}{2} \sqrt{\k^2 - 1}.
    $$
    \item $4\b^2 \s_n^2 \leq 1$ and $4\b^2 \s_1^2 \geq 1$. The optimal $\b$ is achieved at:
    $$
    \frac{1}{2}\left( 1 + \sqrt{1 - 4\b^2 {\s_n^2}} \right) = 2\b^2 \s_1^2 + \b \s_1 \sqrt{4\b^2 \s_1^2 - 1}.
    $$
    The solution is unique since the left is decreasing and the right is increasing. The optimal $\b$ is:
    \begin{equation}
        \b_* = \frac{1}{4\sqrt{2}}\sqrt{\frac{3 \s_1^4 - (\s_1^2 - \s_n^2)^{3/2}\sqrt{9\s_1^2 - \s_n^2} + 6 \s_1^2 \s_n^2 - \s_n^4}{\s_1^4 \s_n^2}}.
    \end{equation}
    The optimal radius satisfies:
    \be
    r_*^2 = \frac{1}{2} + \frac{1}{4\sqrt{2}\s_1^2}\sqrt{(\s_1^2 - \s_n^2)(5 \s_1^2 - \s_n^2 + \sqrt{(\s_1^2 - \s_n^2)(9 \s_1^2 - \s_n^2)})}.
    \en
    This is the optimal solution among the three cases. If $\s_n^2 / \s_1^2$ is small enough we have $r^2 \sim 1 - 1/(3\k^2)$. 
\end{itemize}

\subsubsection{Gauss--Seidel OGD}\label{sec:gs_ogd}
In this subsection, we study Gauss--Seidel OGD and fix $\b_2 = 0$. The characteristic polynomial \eqref{ogd_gs} now reduces to a quadratic polynomial:
$$
\l^2 + (\a^2 \s^2 - 2)\l + 1 - \a \b_1 \s^2 = 0.
$$
For convenience, we reparametrize $\b_1 \to \b/\a$. So, the quadratic polynomial becomes:
$$
\l^2 + (\a^2 \s^2 - 2)\l + 1 - \b \s^2 = 0.
$$
We are doing a min-max optimization $\min_{\a, \b}\max_{\s} r(\a, \b, \s)$, where $r(\a, \b, \s)$ is:
\be
r(\a, \b, \s) = \begin{cases}
\sqrt{1 - \b \s^2} & \a^4 \s^2 < 4(\a^2 - \b)\\
\frac{1}{2}|\a^2 \s^2 - 2| + \frac{1}{2}\sqrt{\a^4 \s^4 - 4 (\a^2 -\b)\s^2} & \a^4 \s^2 \geq 4(\a^2 - \b).
\end{cases}
\en
There are three cases to consider:
\begin{itemize}
    \item $\a^4 \s_1^2 \leq 4(\a^2 - \b)$. We are minimizing $1 - \b \s_n^2$ over $\a$ and $\b$. Optimizing over $\b_1$ gives $\b = \a^2 - \a^4 \s_1^2/4$. Then we minimize over $\a$ and obtain $\a^2 = 2/\s_1^2$. The optimal $\b = 1/\s_1^2$ and the optimal radius is $\sqrt{1 - 1/\k^2}$. 
    \item $\a^4 \s_n^2 > 4(\a^2 - \b)$. Fixing $\a$, the optimal $\b = \a^2 - \a^4 \s_n^2/4$, and we are solving
    $$
    \min_{\a} \max \left\{ \frac{1}{2}|\a^2 \s_1^2 - 2| + \frac{1}{2}\a^2 \sqrt{\s_1^2(\s_1^2-\s_n^2}), \frac{1}{2}|\a^2 \s_n^2 - 2|\right\}.
    $$
    We need to discuss three cases: $\a^2 \s_n^2 > 2$, $\a^2 \s_1^2 < 2$ and $2/\s_1^2 < \a^2 < 2/\s_n^2$. In the first case, the optimal radius is
    $$
    \k^2 - 1 + \k\sqrt{(\k^2-1)}.
    $$
    In the second case, $\a^2 \to 2/\s_1^2$ and the optimal radius is $\sqrt{1 - 1/\k^2}$. In the third case, the optimal radius is also $\sqrt{1 - 1/\k^2}$ minimized at $\a^2 \to 2/\s_1^2$.
    
    \item $\a^4 \s_1^2 > 4(\a^2 - \b)$ and $\a^4 \s_n^2 < 4(\a^2 - \b)$. In this case, we have $\a^2 \s_1^2 < 4$. Otherwise, $r(\a, \b, \s_1) > 1$. We are minimizing over:
    $$
    \max\{\sqrt{1-\b \s_n^2},\, \frac{1}{2}|\a^2 \s_1^2 - 2| + \frac{1}{2}\sqrt{\a^4 \s_1^4 - 4\a^2 \s_1^2 + 4\b \s_1^2}\}.
    $$
    The minimum over $\a$ is achieved at $\a^2 \s_1^2 = 2$, and $\b = 2/(\s_1^2 + \s_n^2)$, this gives $\a = \sqrt{2}/\s_1$ and $\b_1 = \sqrt{2}\s_1/(\s_1^2 + \s_n^2)$. The optimal radius is $r_* = \sqrt{(\k^2 - 1)/(\k^2 + 1)}$. 
 \end{itemize}
    Out of the three cases, the optimal radius is obtained in the third case, where $r\sim 1- 1/{\k^2}$. This is better than Jacobi OGD, but still worse than the optimal EG.

\subsection{Proof of \Cref{hb_opt}: Optimal rate for Gauss--Seidel momentum}\label{app:opt_gs}

In this subsection, we study the optimal spectral radius (convergence rate) of the GS heavy ball method, as we already know that Jacobi HB never converges from \Cref{thm_momentum}. We consider a special case of Gauss--Seidel momentum, the same as in \cite[Theorem 6]{GidelHPHLLM19}: $\b_1 = -1/2$, $\b_2 = 0$. 

\HBopt*

\begin{proof}
The characteristic polynomial becomes:
\be\label{eq:gs_hb_c}
(\l + 1/2)(\l - 1)^2 + \a^2 \s^2 \l^2 = 0,
\en
and we are solving:
\begin{equation}\label{gs_hb_op}
    \min_{\a, \b} \max_{\s \in {\rm Sv}(E)} r(\a, \s),
\end{equation}
with $r(\a, \s)$ the spectral radius of \eqref{eq:gs_hb_c}. The three roots of \eqref{eq:gs_hb_c} are:
\be
\l(\omega, \alpha \sigma) = \frac{1}{6}\left( q(\a \s) - \omega \frac{q(\a \s)^2}{p(\a \s)} - \bar{\omega} p(\a \s) \right),
\en
with
\be
q(t) =  3 - 2t^2, \, p(t) = \left( 54 - 6\sqrt{6}\a \sqrt{27 - 18\a^2 + 4\a^4} - q(t)^3\right)^{1/3}, 
\en
and $\omega \in \{1, -1/2 \pm \sqrt{3}i/2\}$ a cubic root of $1$. $\bar{\omega}$ represents the complex conjugate. We denote $\omega_1 = -1/2 + \sqrt{3}i/2$. Since $p(\sqrt{3/2}) = 0$, we define the functions 
\be
\l(\omega, \sqrt{3/2}) = \lim_{x\to \sqrt{3/2}}\l(\omega, x).
\en
It can be shown that $\l(1, t)$ is a real and monotonically decreasing on $t>0$ with $\l(1, 0) = -1/2$ and $\l(1, \sqrt{2}) = -1$, and $|\l(\omega_1, t)|$ is monotonically decreasing with $|\l(\omega_1, 0)| = 1$. Therefore, the optimal rate $\a$ is the unique positive solution of 
\be
-\l(1, \a \s_1) = |\l(\omega_1, \a \s_n)|,
\en
which we denote as $\a_*$. The analytic solution is difficult but we can find the approximate solution in the feasible domain $0 < \a \s_1 < \sqrt{2}$. Numerically, we find that the optimal convergence rate $r_*$ under the parameter setting $\b_1 = -1/2$, $\b_2 = 0$ satisfies the asymptotic behavior:
\be
r_* = |\l(\omega_1, \a_* \s_n)|\sim 1 - \frac{2}{9\kappa^2},
\en
with $\kappa = \s_1 /\s_n$ the condition number. 
\end{proof}
   
\subsection{Numerical method}\label{appen:num}

We first prove Lemma \ref{rSchur}:
\renewcommand{\thelemma}{4.1}
\begin{lemma}
A polynomial $p(\l)$ is $r$-Schur stable iff $p(r\l)$ is Schur stable.
\end{lemma}

\begin{proof}
Denote $p(\l) = \prod_{i=1}^n (\l -\l_i)$. We have $p(r \l) \propto \prod_{i=1}^n (\l - \l_i/r)$, and:
\begin{equation}
    \forall i\in [n], |\l_i| < r \Longleftrightarrow \forall i\in [n], |\l_i/r| < 1.
\end{equation}
\end{proof}

With Lemma \ref{rSchur} and Corollary \ref{schur_234}, we have the following corollary:
\renewcommand\thecorollary{\thesection.\arabic{corollary}}
\begin{corollary}\label{schur_2_r}
A real quadratic polynomial $\l^2 + a \l + b$ is $r$-Schur stable iff $b < r^2, \, |a| < r + b/r$; A real cubic polynomial $\l^3 + a\l^2 + b \l + c$ is $r$-Schur stable iff $|c| < r^3$, $|ar^2 + c|<r^3+b r$, $br^4 - acr^2 < r^6 - c^2$; A real quartic polynomial $\l^4 + a \l^3 + b \l^2 + c\l + d$ is $r$-Schur stable iff $|c r^5 - ad r^3| < r^8 - d^2$, $|a r^2 +c| < b r+d/r+r^3$, and
$$
b < r^2 +dr^{-2} + r^2\frac{(c r^2 - a d)(a r^2-c)}{(d-r^4)^2}.    
$$
\end{corollary}
\begin{proof}
In Corollary \ref{schur_234}, rescale the coefficients according to Lemma \ref{rSchur}. 
\end{proof}

We can use the corollaries above to find the regions where $r$-Schur stability is possible, i.e., a linear rate of exponent $r$. A simple algorithm might be to start from $r_0 = 1$, find the region $S_0$. Then recursively take $r_{t+1} = s r_{t}$ and find the Schur stable region $S_{t+1}$ inside $S_t$. If the region is empty then stop the search and return $S_t$. $s$ can be taken to be, say, $0.99$. Formally, this algorithm can be described as follows in Algorithm \ref{alg:num}:

\begin{algorithm}[H]\label{alg:num}
\SetAlgoLined
$r_0 = 1$, $t = 0$, $s = 0.99$;

Find the $r_0$-Schur region $S_0$;

 \While{$S_t$ is not empty}{
  $r_{t+1} = sr_t$\;
  Find the $r_{t+1}$-Schur region $S_{t+1}$;\\
  $t = t + 1$;
 }
 \caption{Numerical method for finding the optimal convergence rate}
\end{algorithm}

In this algorithm, Corollary \ref{schur_2_r} can be applied to obtain any $r$-Schur region.  

\section{Supplementary material for Sections \ref{sec:exp} and \ref{sec:conc}}\label{app:exp}

We provide supplementary material for Sections \ref{sec:exp} and \ref{sec:conc}. We first prove that when learning the mean of a Gaussian, WGAN is locally a bilinear game in Appendix \ref{appen:shift}. For mixtures of Gaussians, we provide supplementary experiments about Adam in Appendix \ref{app:adam}. This result implies that in some cases, Jacobi updates are better than GS updates. We further verify this claim in Appendix \ref{jbetter} by showing an example of OGD on bilinear games. Optimizing the spectral radius given a certain singular value is possible numerically, as in Appendix \ref{app:single}.

\subsection{Wasserstein GAN}\label{appen:shift}

Inspired by \cite{daskalakis2017training}, we consider the following WGAN \citep{arjovsky2017wasserstein}:
\begin{equation}
    f(\vphi, \vtheta) = \min_{\vphi} \max_{\vtheta} \E_{\vx\sim \N(\vv, \sigma^2 \mI)} [s(\vtheta^\top \vx)] - \E_{\vz\sim \N({\bf 0}, \sigma^2 \mI)} [s(\vtheta^\top (\vz + \vphi))],
\end{equation}
with $s(x):=1/(1+e^{-x})$ the sigmoid function. We study the local behavior near the saddle point $(\vv, {\bf 0})$, which depends on the Hessian:
$$
\begin{bmatrix}
\n_{\vphi \vphi}^2 & \n_{\vphi \vtheta}^2 \\
\n_{\vtheta \vphi}^2 & \n_{\vtheta \vtheta}^2
\end{bmatrix} = \begin{bmatrix}
-\E_{\vphi}[s''(\vtheta^\top \vz)\vtheta \vtheta^\top] & -\E_{\vphi}[s''(\vtheta^\top \vz)\vtheta \vz^\top + s'(\vtheta^\top \vz)I] \\
(\n_{\vphi \vtheta}^2)^\top & \E_{\vv} [s''(\vtheta^\top \vx)\vx\vx^\top] - \E_{\vphi}[s''(\vtheta^\top \vz)\vz\vz^\top]
\end{bmatrix},
$$

with $\E_{\vv}$ a shorthand for $\E_{\vx\sim \N(\vv, \s^2 \mI)}$ and $\E_{\vphi}$ for $\E_{\vz\sim \N(\vphi, \s^2 \mI)}$. At the saddle point, the Hessian is simplified as:

$$
\begin{bmatrix}
\n_{\vphi \vphi}^2 & \n_{\vphi \vtheta}^2 \\
\n_{\vtheta \vphi}^2 & \n_{\vtheta \vtheta}^2
\end{bmatrix} = 
\begin{bmatrix}
{\bf 0} & -s'(0)\mI \\
-s'(0)\mI & {\bf 0}
\end{bmatrix} = \begin{bmatrix}
{\bf 0} & -\mI/4 \\
-\mI/4 & {\bf 0}
\end{bmatrix}.
$$
Therefore, this WGAN is locally a bilinear game.

\begin{figure}
    \centering
\includegraphics[width=13cm]{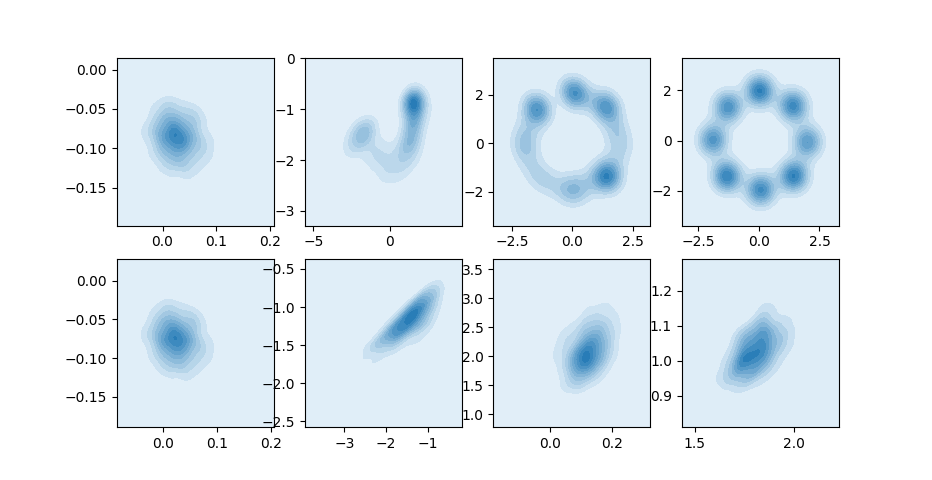}
\caption{Test samples generated from the generator network trained with stochastic Adam. \bb{Top row: } Jacobi updates; \bb{Bottom row:} Gauss--Seidel updates. \bb{Columns} (left to right): epoch 0, 5, 10, 20.}
    \label{fig:adam_j_gs}
\end{figure}

\subsection{Mixtures of Gaussians with Adam}\label{app:adam}

Given the same parameter settings as in Section \ref{sec:exp}, we train the vanilla GAN using Adam, with the step size $\a = 0.0002$, and $\beta_1 = 0.9$, $\beta_2 = 0.999$. As shown in Figure \ref{fig:adam_j_gs}, Jacobi updates converge faster than the corresponding GS updates.

\begin{figure}
    \centering
    \includegraphics[width=0.44\textwidth]{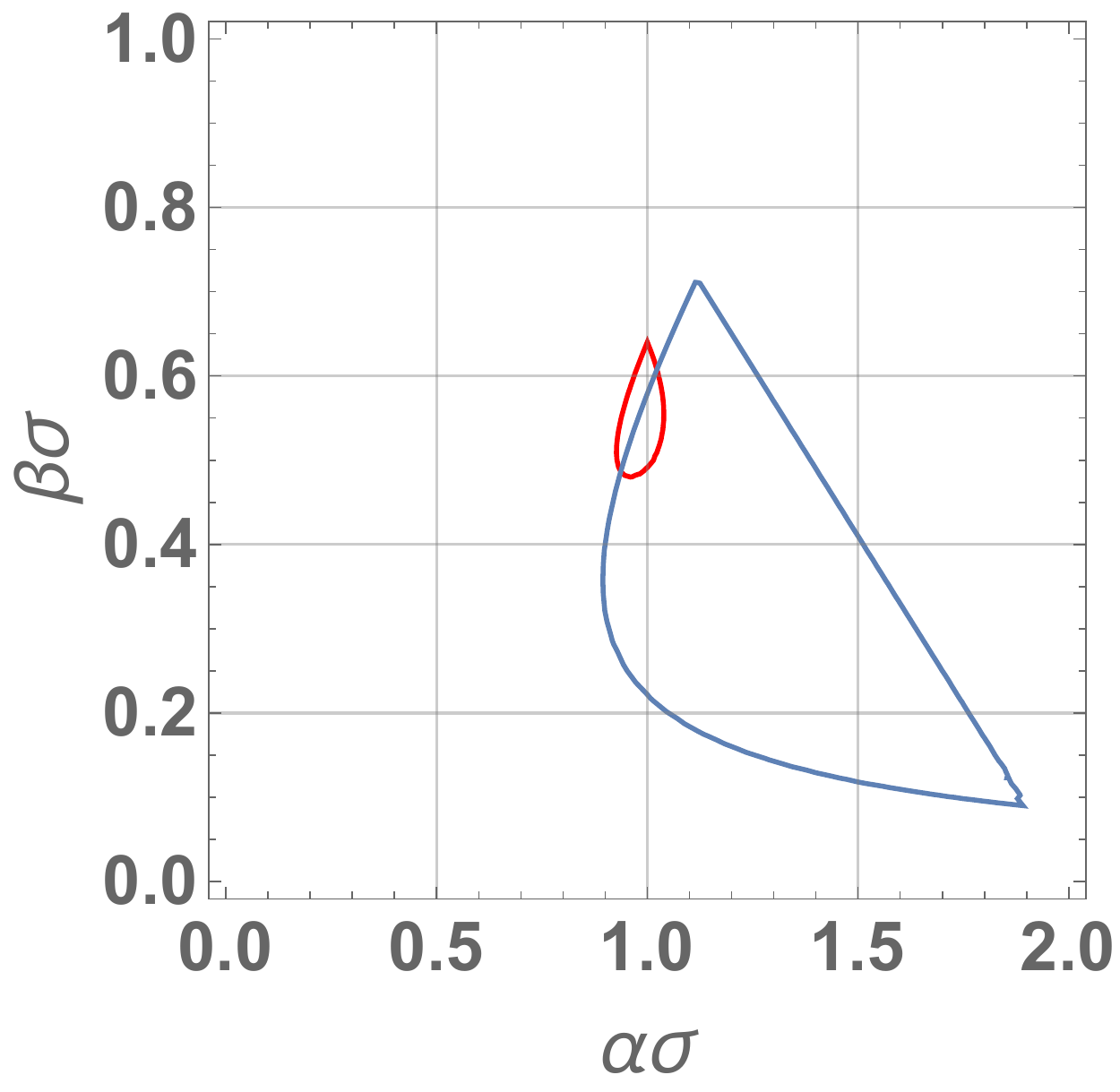}
    \caption{Contour plot of spectral radius equal to $0.8$. The red curve is for the Jacobi polynomial and the blue curve is for the GS polynomial. The GS region is larger but for some parameter settings, Jacobi OGD achieves a faster convergence rate. }
    \label{fig:j-gs-ogd-0.8}
\end{figure}

\subsection{Jacobi updates may converge faster than GS updates}\label{jbetter}
Take $\a = 0.9625$, $\b_1 = \b_2 = \b =  0.5722$, and $\s = 1$, the Jacobi and GS OGD radii are separately $0.790283$ and $0.816572$ (by solving \eqref{ogd_j} and \eqref{ogd_gs}), which means that Jacobi OGD has better performance for this setting of parameters. A more intuitive picture is given as Figure \ref{fig:j-gs-ogd-0.8}, where we take $\b_1 = \b_2 = \b$.

\subsection{Single singular value}\label{app:single}
We minimize $r(\vtheta, \s)$ for a given singular value numerically. WLOG, we take $\s = 1$, since we can rescale parameters to obtain other values of $\s$.
We implement grid search for all the parameters within the range $[-2, 2]$ and step size $0.05$. For the step size $\a$, we take it to be positive. We use $\{a, b, s\}$ as a shorthand for $\{a, a + s, a+2s, \dots, b\}$.
\begin{itemize}
    \item We first numerically solve the characteristic polynomial for Jacobi OGD \eqref{ogd_j}, fixing $\a_1 = \a_2 = \a$ with scaling symmetry. With $\a \in \{0,2, 0.05\}$, $\b_i \in \{-2, 2, 0.05\}$, the best parameter setting is $\a = 0.7$, $\b_1 = 0.1$ and $\b_2 = 0.6$. $\b_1$ and $\b_2$ can be switched. The optimal radius is $0.6$.
    \item We also numerically solve the characteristic polynomial for Gauss--Seidel OGD \eqref{ogd_gs}, fixing $\a_1 = \a_2 = \a$ with scaling symmetry. With $\a \in \{0,2, 0.05\}$, $\b_i \in \{-2, 2, 0.05\}$, the best parameter setting is $\a = 1.4$, $\b_1 = 0.7$ and $\b_2 = 0$. $\b_1$ and $\b_2$ can be switched. The optimal rate is $1/(5\sqrt{2})$. This rate can be further improved to be zero where $\a = \sqrt{2}$, $\b_1 = 1/\sqrt{2}$ and $\b_2 = 0$.
    \item Finally, we numerically solve the polynomial for Gauss--Seidel momentum \eqref{hb_gs}, with the same grid. The optimal parameter choice is $\a = 1.8$, $\b_1 = -0.1$ and $\b_2 = -0.05$. $\b_1$ and $\b_2$ can be switched. The optimal rate is $0.5$.
\end{itemize}

\section{Splitting Method}\label{split}

In this appendix, we interpret the gradient-based algorithms (except PP) we have studied in this paper as splitting methods \citep{Saad03}, for both Jacobi and Gauss--Seidel updates. By doing this, one can understand our algorithms better in the context of numerical linear algebra and compare our results in Section \ref{schur_ana} with the Stein--Rosenberg theorem. 
\subsection{Jacobi updates}\label{append:a1_j}
From \eqref{sp}, finding a saddle point is equivalent to solving:
\begin{equation}\label{saddle_point_eq}
    \mS\vz :=\begin{bmatrix}
    {\bf 0} & \mE \\
    -\mE^\top & {\bf 0}
    \end{bmatrix}
    \begin{bmatrix}
    \vx \\
    \vy
    \end{bmatrix} = 
        \begin{bmatrix}
    -\vb \\
    \vc
    \end{bmatrix} =: \vd.
\end{equation}

Now, we try to understand the Jacobi algorithms using splitting method. For GD and EG, the method splits $\mS$ into $\mM - \mN$ and solve
\begin{equation}
    \vz_{t+1} = \mM^{-1}\mN \vz_t + \mM^{-1}\vd.
\end{equation}
For GD, we can obtain that:
\begin{equation}
    \mM = \begin{bmatrix} 
    \a_1^{-1}\mI & {\bf 0} \\
    {\bf 0} & \a_2^{-1}\mI 
    \end{bmatrix},\,    
    \mN = \begin{bmatrix} 
    \a_1^{-1}\mI & -\mE \\
    \mE^\top & \a_2^{-1}\mI 
    \end{bmatrix}.
\end{equation}
For EG, we need to compute an inverse:
\begin{equation}
    \mM^{-1} = \begin{bmatrix} 
    \a_1 \mI  & -\b_1 \mE \\
    \b_2 \mE^\top & \a_2 \mI
    \end{bmatrix}, \, \mN = \mM - \mS.
\end{equation}
Given $\det(\a_1 \a_2 \mI + \b_1 \b_2 \mE\mE^\top)\neq 0$, the inverse always exists. 

The splitting method can also work for second-step methods, such as OGD and momentum. We split $\mS = \mM - \mN - \mP$ and solve:
\begin{equation}
        \vz_{t+1} = \mM^{-1}\mN \vz_t + \mM^{-1}\mP \vz_{t-1} + \mM^{-1}\vd.
\end{equation}

For OGD, we have:
\begin{equation}
    \mM = \begin{bmatrix} 
    \frac{\mI}{\a_1-\b_1} & {\bf 0} \\
    {\bf 0} & \frac{\mI}{\a_2-\b_2}
    \end{bmatrix},  \, 
    \mN = \begin{bmatrix} 
    \frac{\mI}{\a_1-\b_1} & -\frac{\a_1 \mE}{\a_1 - \b_1} \\ \\
    \frac{\a_2 \mE^\top}{\a_2 - \b_2} & \frac{\mI}{\a_2-\b_2}
    \end{bmatrix},\,
    \mP = \begin{bmatrix} 
    {\bf 0} & \frac{\b_1 \mE}{\a_1 - \b_1} \\
    -\frac{\b_2 \mE^\top}{\a_2 - \b_2} & {\bf 0}
    \end{bmatrix}.
\end{equation}
For the momentum method, we can write:
\begin{equation}
    \mM = \begin{bmatrix} 
    \a_1^{-1}\mI & {\bf 0} \\
    {\bf 0} & \a_2^{-1}\mI 
    \end{bmatrix},\, 
    \mN = \begin{bmatrix}
     \frac{1+\b_1}{\a_1}\mI & - \mE \\
      \mE^\top & \frac{1+\b_2}{\a_2} \mI
    \end{bmatrix},\, \mP =  \begin{bmatrix}
      -\frac{\b_1}{\a_1} \mI & {\bf 0} \\
      {\bf 0} & -\frac{\b_2}{\a_2} \mI 
    \end{bmatrix}.
\end{equation}

\subsection{Gauss--Seidel updates}\label{append:a1_gs}
 Now, we try to understand the GS algorithms using splitting method. For GD and EG, the method splits $\mS$ into $\mM - \mN$ and solve
\begin{equation}
    \vz_{t+1} = \mM^{-1}\mN \vz_t + \mM^{-1}\vd.
\end{equation}
For GD, we can obtain that:
\begin{equation}
    \mM = \begin{bmatrix} 
    \a_1^{-1}\mI & {\bf 0} \\
    -\mE^\top & \a_2^{-1}\mI 
    \end{bmatrix},\,    
    \mN = \begin{bmatrix} 
    \a_1^{-1}\mI & -\mE \\
    {\bf 0} & \a_2^{-1}\mI 
    \end{bmatrix}.
\end{equation}
For EG, we need to compute an inverse:
\begin{equation}
    \mM^{-1} = \begin{bmatrix} 
    \a_1 \mI  & -\b_1 \mE \\
    (\b_2 + \a_1 \a_2) \mE^\top & \a_2 (\mI - \b_1 \mE^\top \mE)
    \end{bmatrix}, \, \mN = \mM - \mS.
\end{equation}

The splitting method can also work for second-step methods, such as OGD and momentum. We split $\mS = \mM - \mN - \mP$ and solve:
\begin{equation}
        \vz_{t+1} = \mM^{-1}\mN \vz_t + \mM^{-1}\mP \vz_{t-1} + \mM^{-1}\vd.
\end{equation}

For OGD, we obtain:
\begin{equation}
    \mM = \begin{bmatrix} 
    \frac{\mI}{\a_1-\b_1} & {\bf 0} \\ \\
    -\frac{\a_2 \mE^\top}{\a_2 - \b_2} & \frac{\mI}{\a_2-\b_2}
    \end{bmatrix},  \, 
    \mN = \begin{bmatrix} 
    \frac{\mI}{\a_1-\b_1} & -\frac{\a_1 \mE}{\a_1 - \b_1} \\ \\
    -\frac{\b_2 \mE^\top}{\a_2 - \b_2} & \frac{\mI}{\a_2-\b_2}
    \end{bmatrix},\,
    \mP = \begin{bmatrix} 
    {\bf 0} & \frac{\b_1 \mE}{\a_1 - \b_1} \\ \\
    {\bf 0} & {\bf 0}
    \end{bmatrix}.
\end{equation}
For the momentum method, we can write:
\begin{equation}
    \mM = \begin{bmatrix} 
    \a_1^{-1}\mI & {\bf 0} \\
    -\mE^\top & \a_2^{-1}\mI 
    \end{bmatrix},\, 
    \mN = \begin{bmatrix}
     \frac{1+\b_1}{\a_1}\mI & - \mE \\
      {\bf 0} & \frac{1+\b_2}{\a_2} \mI
    \end{bmatrix},\, \mP =  \begin{bmatrix}
      -\frac{\b_1}{\a_1} \mI & {\bf 0} \\
      {\bf 0} & -\frac{\b_2}{\a_2} \mI 
    \end{bmatrix}.
\end{equation}

\section{Singular bilinear games}\label{app:singular}
In this paper we considered the bilinear game when $\mE$ is a non-singular square matrix for simplicity. Now let us study the general case where $\mE\in \R^{m\times n}$. As stated in Section \ref{prem}, saddle points exist iff 
\be
\vb\in \rr(\mE), \, \vc\in \rr(\mE^\top).
\en
Assume $\vb = \mE \vb'$, $\vc = \mE^\top \vc'$. One can shift the origin of $\vx$ and $\vy$: $\vx \to \vx - \vb'$, $\vy \to \vy - \vc'$, such that the linear terms cancel out. Therefore, the min-max optimization problem becomes: 
\be
\min_{\vx\in \R^m}\max_{\vy\in \R^n} \vx^\top \mE \vy.
\en
The set of saddle points is:
\be
\{(\vx, \vy)|\vy\in \N(\mE), \vx\in \N(\mE^\top)\}.
\en
For all the first-order algorithms we study in this paper, $\x{t}\in \x{0} + \rr(\mE)$ and $\y{t}\in \y{0} + \rr(\mE^\top)$. Since for any matrix $\mX\in \R^{p\times q}$, $\rr(\mX)\oplus \N(\mX^\top) = \R^p$, if the algorithm converges to a saddle point, then this saddle point is uniquely defined by the initialization:
\be
\vx^* = P_\mE^\perp\x{0}, \, \vy^* = P_{\mE^\top}^\perp\y{0},
\en
where
\be
P_\mX^\perp := \mI - \mX^\dagger \mX, 
\en
is the orthogonal projection operator onto the null space of $\mX$, and $\mX^\dag$ denotes the Moore--Penrose pseudoinverse. Therefore, the convergence to the saddle point is described by the distances of $\x{t}$ and $\y{t}$ to the null spaces $\N(\mE^\top)$ and $\N(\mE)$. We consider the following measure:
\be\label{eq:app_dis}
\Delta_t^2 = ||\mE^\dagger \mE\y{t}||^2 + ||\mE\mE^\dagger \x{t}||^2,
\en
as the Euclidean distance of $\z{t} = (\x{t}, \y{t})$ to the space of saddle points $\N(\mE^\top)\times \N(\mE)$. Consider the singular value decomposition of $\mE$:
\be
\mE = \mU \begin{bmatrix} 
\mSigma_r & \zero \\
\zero & \zero
\end{bmatrix} \mV^\top,
\en
with $\mSigma_r\in \R^{r\times r}$ diagonal and non-singular. Define:
\be
\vv^{(t)} = \mV^\top \y{t}, \, \vu^{(t)} = \mU^\top \x{t},
\en
and \eqref{eq:app_dis} becomes:
\be
\Delta_t^2 = ||\vv^{(t)}_r||^2 + ||\vu^{(t)}_r||^2,
\en
with $\vv_r$ denoting the sub-vector with the first $r$ elements of $\vv$. Hence, the convergence of the bilinear game with a singular matrix $\mE$ reduces to the convergence of  the bilinear game with a non-singular matrix $\mSigma_r$, and all our previous analysis still holds. 
\end{document}